\documentclass[11pt]{article}

\usepackage{natbib}
\usepackage{geometry}
\usepackage{times}
\usepackage{latexsym}
\usepackage{amsmath}

\usepackage[T1]{fontenc}

\usepackage[utf8]{inputenc}

\usepackage{microtype}

\usepackage{inconsolata}

\usepackage{xcolor}
\usepackage{soul}

\usepackage{multirow}
\usepackage{booktabs}
\usepackage{adjustbox}

\usepackage{linguex}
\usepackage{relsize}

\usepackage{caption}
\usepackage{subcaption}
\usepackage{float}

\usepackage{todonotes}

\usepackage[]{titlesec}
\titlespacing*{\paragraph}{0pt}{1ex}{1ex}
\usepackage{enumitem}

\usepackage{hyperref}
\hypersetup{
    colorlinks=true,
    citecolor=blue,
    linkcolor=blue,
    urlcolor=blue,
    breaklinks=true
}
\usepackage{url}

\usepackage[nameinlink,capitalize]{cleveref}

\newcommand{\keywords}[1]{\par\vspace{0.5em}\noindent\textbf{Keywords:} #1}

\title{
Large Language Models and Impossible Language Acquisition: ``False Promise'' or an Overturn of our Current Perspective towards AI
}

\author{
Ziyan Wang$^{1}$ \\
\textbf{Longlong Ma}$^{2\dagger}$ \\[1ex]
${}^{1}$New Talent Academy, Beijing \\
${}^{2}$Institute of Software, University of Chinese Academy of Sciences \\[1ex]
\texttt{xieguaiwu@163.com}, \texttt{longlong@iscas.ac.cn}
}

\begin{document}

\setlength{\Exlabelwidth}{1em}
\setlength{\Exlabelsep}{0.7em}
\setlength{\SubExleftmargin}{1.3em}
\setlength{\Extopsep}{2pt}

\maketitle

\begin{abstract}
In Chomsky's provocative critique ``The False Promise of CHATGPT,'' Large Language Models (LLMs) are characterized as mere pattern predictors that do not acquire languages via intrinsic causal and self-correction structures like humans, therefore are not able to distinguish impossible languages. It stands as a representative in a fundamental challenge to the intellectual foundations of AI, for it integrally synthesizes major issues in methodologies within LLMs and possesses an iconic \textit{a priori} rationalist perspective. We examine this famous critique from both the perspective in pre-existing literature of linguistics and psychology as well as a research based on an experiment inquiring into the capacity of learning both possible and impossible languages among LLMs. We constructed a set of syntactically impossible languages by applying certain transformations to English. These include reversing whole sentences, and adding negation based on word-count parity. Two rounds of controlled experiments were each conducted on GPT-2 small models and long short-term memory (LSTM) models. Descriptive analysis of single-run training trajectories shows that GPT-2 small models exhibit lower final loss, faster convergence, and lower perplexity on natural language compared to impossible language conditions, with the reversed condition showing the largest departure (loss ratios up to 2.25$\times$ natural). LSTM models, by contrast, show minimal differences across conditions. Given the single-run nature of our experiments ($n=1$ per condition), we report descriptive comparisons and caution that formal statistical inference is precluded. Based on theoretical analysis and descriptive empirical findings, we propose a new vision within Chomsky's theory towards LLMs, and a shift of theoretical paradigm outside Chomsky, from his ``rationalist-romantics'' paradigm to functionalism and empiricism in LLMs research.
\end{abstract}


\section{Introduction}

The rapid recent developments in Large Language Models (LLMs) have intrigued debates across multiple disciplines, from studies that seek to study mentality as an abstract cognitive subject, such as philosophy of mind and cognitive science, to internal disagreements among different paradigms within AI. For the former, the skeptical criticism from \citet{chomsky2023nyt} based on his theories, such as Universal Grammar (UG) and ``Poverty of the Stimulus'' (PoS) in language acquisition, is especially notable, given the long-standing dominant status of Chomsky's framework in various subjects other than linguistics that focus on language learning and syntactic organization. As a result, the obvious part of his critique has often been taken as hard fact, even though it relies on unascertainable assumptions. A careful examination therefore, requires investigations with respect to two parallel tracks.

\subsection{The Partition in Chomsky's Doctrine}

One must recognize that there is an essential distinction not only within the Chomskyan linguistics, but also inside his general opinion related to the philosophy of mind: the division between linguistic competence (the explanatory hypothesis that attempts to depict language acquisition universally) and syntactic analysis (the formal modeling of grammatical structures). His criticism towards LLMs is mainly derived from the part of linguistic competence in his theory that describes the origin and nature of human language faculty from an external view, whose methodological foundation differs entirely from the mathematical formalization in the part of syntactic analysis, where Chomsky gained much of his influence. The core arguments of the former are philosophical and metaphysical, involving questions that cannot be solved solely through empirical science, but rather remain in debate. This divergence is often neglected in the public reception of the criticism, but it is the crux of Chomsky's line of reasoning.

The realization that there are non-negligible gaps between the achievements in formal syntactic analysis and the universal rationalist claims on linguistic competence provides counter-evidence against Chomsky's key arguments. Attempts to support the totality of the Chomskyan paradigm on mentality in virtue of the authority from the syntactic analytical framework are unreliable, and moreover, the fact that Chomsky's linguistic theories have little contribution to the overall achievement in LLMs could suggest a necessity to reexamine the pervaded perspective related to language acquisition in AI.

\subsection{Inspecting the Logical Weakness}

In order to reevaluate the Chomskyan perspective sufficiently, the investigation is conducted on two interconnected aspects, demanding both a philosophical view and empirical experiments, because of the logical necessity for those discoveries on a shift of paradigm to construct theoretical judgments of principles besides objective evidence.

\subsubsection{In Theoretical Aspect}

On top of all, the philosophical basis of Chomsky's specific criticisms includes two core arguments: (a) LLMs are merely instruments for descriptions and predictions based on vast data, contrasting to the PoS in the learning process of humans, without acknowledging the underlying causal relations among events or grammatical principles in languages; (b) LLMs lack an \textit{a priori} ``language organ'' \citep{chomsky2009}, and do not possess true intelligence in language acquisition, for they are naturally pattern predictors. Therefore, LLMs are unable to judge whether a language is intrinsically possible or not, an ability Chomsky deems natural in humans.

In this fundamental discussion, we engage with classic literature from both developmental psychology and philosophy, which both offer alternative understandings of Chomsky's questions, and furthermore, new theoretical architectures. For instance, we have drawn upon Jean Piaget's constructivist theory of cognitive development \citep{piaget1970}, which describes the staging progress of human cognition, engaging with the adaptation to unfamiliar experience.

From philosophy, we introduce Gilbert Ryle's classic work \textit{The Concept of Mind} to support our disproof. In his critique of Cartesian dualism, Ryle proposes a practical vision on intellectual operations, which relies on the ``intelligent practice'' that concerns external behaviors \citep{ryle2009}, whereas his dissent to ``the dogma of the ghost in the machine'' \citep{ryle2009} directly challenges the metaphysical core of Chomsky's ``language organ.'' Ryle's framework shares a common ground with the empirical paradigm in science, which contributes largely to our concerns on paradigm shift. It claims that the indicators to identify intellectual capacity should be behaviors, rather than different metaphysical blueprints.

Although being widely accepted, the PoS Argument (PoSA) that props the Chomskyan UG is not flawless even in pure reasoning. A recent literature review challenges the establishment of PoSA, and displays the dilemma for it to back up linguistic innatism \citep{skidelsky2016}. Whereas studies concerning the learnability of language in statistical learning \citep{lewis2001,perfors2011} have shown AIs' real performance in language acquisition as learners with similarities to human, rather than the \textit{prima facie} that regard them as parroting programs.

\subsubsection{Our Path Forward}

The philosophical research indicates an apparent shift in theoretical perspective: from the Chomskyan ``rationalist-romantics'' paradigm \citep{chomsky2009} towards a functionalist and empiricist one, to accommodate the future development of LLMs. Such a shift would revisit approaches, like those of B. F. Skinner and K. Halliday. Skinner's experiment-oriented methodology points to the involvement of common scientific logic in the technical practice of AI, while Halliday's developmental thought in functional linguistics shows a divergent view that straightforwardly regards ``both language use and context center stage in linguistics,'' rejecting studies that depend on ``the whims of a single individual -- Saussurean \textit{sujet parlant} nor the Chomskyan `ideal native speaker','' but the scientific linguistics that establishes the nature of language ``on the exchanges of meaning between ordinary speakers as participants in some concerted social activity'' instead \citep{halliday2009}. Both of them treat language as a learnable skill shaped by environmental interaction.

\subsubsection{A Clarification of our Intention}

Although it appears that we are standing on the opposite side from the Chomskyan school, we are not supporting an environmentalism based on the Anglo-American empiricist tradition, which believes that ``the mind results from a few simple operations of association on the basis of contiguity,'' and for humans alone any intellectual barrier associated with language acquisition can be overcome by learning by whatever means \citep{chomsky1973}. We are not contesting theories of the unique biological structures that inherently empower humans to acquire languages. The biological view on environment and experience in \citet{chomsky2005} could represent a consensus between us and Chomsky. Though we suggest that there is comparability between the learning process in humans and in LLMs, the essential differences between them apparently demonstrate the impossibility of transplanting every conclusion from machine learning to humans, regardless of the external survey on LLMs. The only kinds of comparability between humans and LLMs we can accept are those shown in intellectual behaviors, which do not contrast with their neurological incommensurability, for LLMs possess no biological substrate, but hold comprehensive experience. This study only intends to focus on the problems of Chomsky's critiques within practical considerations for the future development of machine learning, and what his limitations have revealed in his cognitive theory.

\subsubsection{Preparation and Methodology of the Experiment}

The aforementioned theoretical juxtaposition still requires fundamental evidence to internally reconsider Chomsky's criticism that can be empirically measured. Such supporting evidence is already concluded by a few pre-existing studies, which show the universality of related experience, enhancing its influence on the general methodological choices. This evidence concerns the differential learning efficiency of LLMs when trained on impossible languages. In Chomsky's original critique, the core argument (a) is directly relevant to the issue. It is claimed that LLMs are merely pattern predictors, unable to distinguish an impossible language, but, on the contrary, both the experiments that we performed and those from \citet{kallini2024} suggest differential performance of LLMs when acquiring impossible languages compared to their capacity in learning natural languages.

The experiment that takes place in our paper also intends to examine the claim that LLMs are equally adequate to learn both possible and impossible languages, judged by the Chomskyan studies on syntactic structures. \citet{kallini2024} have provided a mature basic design to evaluate and quantify the performance of LLMs in natural languages and impossible languages. Our experiment is partly based on that method, but also based on our original improvements: firstly, we chose not only the BabyLM dataset \citep{warstadt2023} as the corpus for our training, for it is not always practical to implement impossible grammatical changes on complex sentences while leaving the linguistically unnatural characteristics apparent for LLMs. It is essential to avoid difficulties such as word segmentation, long-sentence structures, and other factors that could disturb LLMs' acquisition process, and exclude variables other than the possibility of each language itself. Thus, we used an automatic mechanism to generate simple SVO sentences for every grammatical transformation, which is also beneficial to strengthen the universality and validity of our findings in a larger variety of datasets other than BabyLM, as well as in making it easier for researchers who lack hardware resources like ourselves. Secondly, in the precise choices of indicators for the effectiveness of learning processes, we applied the data collection of both perplexity and loss value, whereas \citet{kallini2024} only focused on perplexity. In this way, our experiment can be more objective in the deduction of conclusions.

\begin{figure}[ht]
    \centering
    \includegraphics[width=0.65\textwidth]{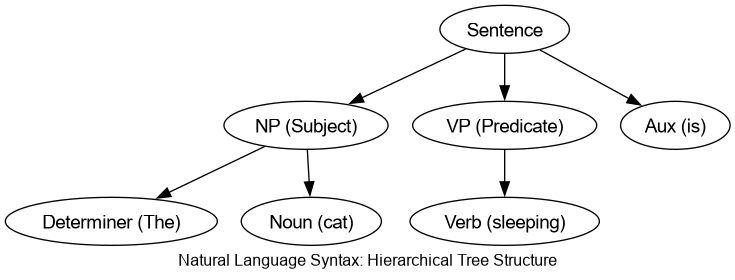}
    \caption{Natural language operates on hierarchical tree structures}
    \label{fig:natural-structure}
\end{figure}

\begin{figure}[ht]
    \centering
    \includegraphics[width=0.65\textwidth]{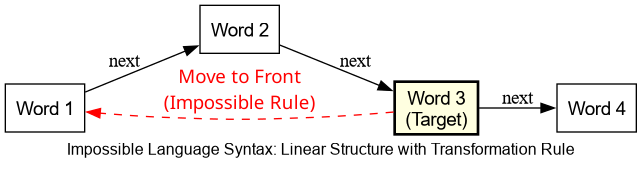}
    \caption{Impossible languages rely on rigid linear positioning}
    \label{fig:impossible-structure}
\end{figure}

There could be some possible questions when it comes to the specific practice of the experiment, referring to the construction of impossible languages: how are they actually created, and in what way do they differ from natural human languages? We employed rule-based transformations to English, where the original language itself does not intervene in acquisition process, hence does not count as an influential factor in the experiment. The transformations are as follows: whole-sentence reversal, and inserting negation governed by word-count parity. They are undoubtedly artificial in Chomsky's syntactic theory, for they all rely on linear positioning, which is structurally against the rules of natural language (Figure~\ref{fig:natural-structure} \& Figure~\ref{fig:impossible-structure}). Of course, they are not the only possible choices, but they are more connected to the structural component compared to those in \citet{kallini2024} that selected Random Word Shuffles as one of the impossible transformations. Considering Chomsky's attention to the well-ordered structure of natural language, our choices in the experiment could be more relevant in supporting our argument.

\subsubsection{Our Empirical Assessment}

In our experiment, three GPT-2 models and LSTM models are separately trained:

\begin{itemize}
    \item \textbf{Group 1: Natural Language Group} (control group), where the model was trained on the original dataset.
    \item \textbf{Group 2: Reversed Group} (experimental group 1), where the model was trained on a dataset with all sentences reversed.
    \item \textbf{Group 3: Parity Negation Group} (experimental group 2), where the model was trained on a dataset that is applied on parity negation transformation, which inserts negation in different positions based on word-count parity.
\end{itemize}

At first, our experiment was designed to verify Chomsky's opinion on LLMs. The primal hypothesis was that the models of each kind trained on both possible and impossible languages can achieve similar average perplexity, which would suggest that LLMs are incapable of identifying natural and unnatural languages. However, our descriptive results suggest a different picture: GPT-2 models trained on natural language achieved the lowest loss values and perplexities throughout training, while the reversed condition consistently showed the largest departure. The parity negation condition showed mixed results---clearly worse than natural on the small dataset, but nearly indistinguishable on BabyLM. Two rounds of experiments conducted on two datasets yielded a consistent descriptive pattern: GPT-2 models learn natural languages more efficiently than impossible languages under most transformations, while LSTM models show minimal differences across conditions. We emphasize that these are single-run observations ($n=1$ per condition) requiring replication for statistical validation.\footnote{The code for the experiment can be found at \url{https://github.com/xieguaiwu/LLMs-and-impossible-language-acquisition}}

\section{Background and Related Works}

\subsection{An Analysis for Chomsky's Critique}

\subsubsection{Line of Reasoning}

In \citet{chomsky2023nyt}, his critique is based on two crucial theses:

\begin{enumerate}
    \item[(a)] According to the PoSA, there exists a natural physiological structure in the human brain that endows the creative ability to humans. On the opposite, LLMs are nothing but advanced parrots, which can serve quite well in certain tasks by imitating human brains from vast data.
    \item[(b)] Since LLMs do not naturally have this prior structure, they can never possess true intelligence regardless of how much they learn.
\end{enumerate}

The propositions themselves are logical inferences from Chomsky's nativism, where (a) uses PoSA to certify the physical property of language. Before investigating PoSA with the Chomskyan paradigm, we need to examine his ratiocination about the linguistic apriority at first.

In \citet{chomsky1988}, after claiming that language acquisition is a physical mechanism, Chomsky then clarifies its nature of innateness: ``If a creature has the capacity to perform certain tasks well, then these very capacities will lead to failure in some other tasks.'' Hence, he deduces the fundamental nature of language competence: ``In the case of language the language faculty, a physical mechanism in the sense already explained, has certain definite properties... These properties permit the human mind to acquire a language of a specific type... The same properties exclude other possible languages as `unlearnable' by the language faculty.'' \citep{chomsky1988}

Undeniably, language competence must have a physiological and physical foundation. However, limited empirical knowledge is not enough evidence to consequentially describe this foundation as ``a `language acquisition device' that takes experience as `input' and gives the language as an `output'.''' \citep{chomsky2000} Apparently, any insight we can have on the mind is bounded on observable phenomenon, from nervous activities to verbal communications. Numerous mind philosophers have proposed divergent theories with different tendencies as well as searchable logical discontinuities, due to a contradiction between the desire for a universal description, and the chaotic reality that leaves little possibility to experientially find one, where nothing can make one theory truer than the others. When logical discontinuities in a seemingly complete theory such as the one from Chomsky are found, it is likely that those discontinuities arise from the conceptualization of observations, which itself structuralizes more that can be supported by the evidence.

In Chomsky's example, he structuralizes a pure and absolute metaphysical innateness from early children's verbal behavior \citep{chomsky1973,chomsky1988,chomsky2000,chomsky2002,chomsky2009}, and then reinforces his criticism on LLMs with such conceptualization that lacks sufficient empirical support in the context of machine learning, despite that there are also direct observations that can support alternative views as well, which would be an intuitive and experiential support without enough statistical effectiveness, just like that in Chomsky works. Same insight on Chomsky's methodology can also be seen in \citet{hintikka1999}, where it was regarded as the origin of the use of intuitions in philosophical argumentation after the mid-1960s.

\subsubsection{External Analysis}

Philosophically, the problem with Chomsky's argument is similar to what Quine criticizes logical positivism. Quine targets the insufficient distinction between analytic and synthetic statement, especially the one drawn by Carnap's artificial language, which is: ``a distinction to be drawn at all is an unempirical dogma of empiricists, a metaphysical article of faith.'' \citep{quine1951} In the same meaning, Chomsky also draws such a line in the middle of language faculty and language performance, where he thereby divides his theory into two parts -- language competence and syntactic analysis. In reality, any hint of a language faculty that is stored inside a ``language organ'' is given by language performance.

We are coming to a branch point, an essential divergence of nativism and functionalism, and that of a metaphysical rationalism and an experiment-oriented pragmatism in science. The former used to be a leap from Skinner's behaviorist linguistic perspective, which was seen as an inheritor of mentalism with only terminological revisions but still remain invalid \citep{chomsky1959}, because of both the lack of experimental observations and its intention to overwhelm empirical deficiency with structuralization. At this moment, perhaps Chomsky's view is also a theory as such.

In \citet{skidelsky2016}, the PoSA was typologically examined: as an inference to the best explanation, PoSA can only reliably establish domain-specificity that language acquisition requires linguistically-specific principles, but not suffice to establish innatism in linguistic technicality nor UG. Therefore, a dilemma has shown: ``The problem is that the arguments proposed in the literature turn the innateness hypothesis into what seems to be a feeble empirical claim, when considered optimistically, or an armchair piece of knowledge, when considered pessimistically.'' \citep{skidelsky2016}

As a matter of fact, specific conclusions from quantitative experiments have contradicted thesis (a), whereas (b) still remains unfalsifiable. Three experiments conducted by a renowned study \citep{kallini2024} to evaluate the difference of GPT-2 models' capacities in learning possible language and multiple impossible languages, where their performances worsen as the language type gets more impossible. Conclusions from the study were on the basis of the BabyLM dataset.

A classic study also supports the learnability of language in statistical learning. It was found in \citet{lewis2001} that even simple recurrent networks (SRNs) can capture the inner structure of relative clauses, suggesting the emergence of syntactic structural understanding. On the same level, \citet{perfors2011} indicated with Bayesian architecture that a learner with only general inductive abilities is already able to identify hierarchical phrase structures. Furthermore, they clarified the problem of PoSA in practice: what should be focused on is ``what kind of knowledge must be assumed as an innate constraint on the learner's inductive hypotheses, rather than on what kind of representational machinery must be available to the learner.'' \citep{perfors2011}

\subsubsection{Synthetical Understanding}

This knowledge leads us to a critical reevaluation towards the nativist refuse on Turing's test. Not only do LLMs have ``a capacity to produce boundless numbers of sentences and understand and interpret them'' \citep{chomsky2009}, but evidence also suggests that statistical learning mechanism can acquire languages even in a manner comparable to human cognitive processes. Consequently, a linguistic theory that tends to become ``a mathematical description of the operations that these brain states and events carry out and the mental sensory states they can assume, as Newton's inverse square law'' \citep{chomsky2009} cannot overlook the progress in AI. Indeed, Chomsky's works on the foundation of language organization has built a complete and formalized academic edifice, but its theoretical evolution over decades has often relied on rationalist introspection, rather than the iterative, data-driven feedback loops in the methodology of empirical natural science. This fact reveals what is left to be done: a developmental and functional paradigm that can apply empirical experiments on the basis of the Chomskyan framework. For this objective, it is essential to work on a new perspective on LLMs.

\subsection{Experimental Approach}

\subsubsection{Past Conclusions}

Besides Chomsky's original critique, other recent representational literature have also claimed the equality of learning both possible and impossible language for LLMs \citep{bolhuis2024,moro2023}. However, in \citet{kallini2024}, this thesis is empirically questioned under the advanced transformer architecture. Comparing with previous literatures that validate it \citep{mitchell2020}, the progress of framework is indicated to be the crux of such difference. Contacting that to \citet{linzen2016}, whose conclusions have reflected the grammatical sensibility in RNNs like \citet{lewis2001} and \citet{perfors2011}, but on the other hand, reveal RNN's ineffectiveness in difficult cases, indicating the influence of training framework. The notable divergence between the experiments that we conduct on GPT-2 and those that are conducted on LSTM is also a case in point. With such knowledge, a pure empiricist view like Locke's tabula rasa theory on language acquisition is empirically repudiated \citep{kodner2023}.

The decisive role of preset conditions, from learning mechanisms to various constraints, are essential in a functional paradigm.

\subsubsection{Model Diversity}

While \citet{kallini2024} reconsiders the empirical validity of Chomsky's argument, and thereby, creates a solid methodological foundation for our research, we investigate the role of neural network architecture as a factor in language acquisition by enhancing model diversity, applying LSTM model as an example of RNN models. LSTM model's performance contrast with GPT2 small model's not only in our experiment, but also in impossible language experiments conducted by \citet{gulordava2018} and \citet{mitchell2020} on LSTM models, versus that done by \citet{kallini2024} on GPT2 small models. As a secondary experimental subject, LSTM model allows for a more rigorous assessment, and a architectural perspective as well.

\subsubsection{Impossible Language Training}

\citet{moro2016} and \citet{moro2023} define impossible language as non-hierarchical and linear (see Figure~\ref{fig:impossible-structure}), radically divergent from the recursive property in Chomsky's description \citep{chomsky1957,chomsky1965}. Moreover, \citet{moro2003} points to the existence of a clinical difference for humans to learn possible and impossible languages, suggesting that Broca's area, an essential part of the brain to fulfill speech production functions, ``is progressively disengaged'' when dealing with unreal grammars. Andrea Moro's other important works are also dedicated to discussing the neurological foundation of impossible languages identification: in \citet{moro2000}, the tripartite structure of language \citep{chomsky1957,chomsky1965,chomsky2014} is discovered to have a corresponding distributed neural network implementation, which relates to the subjects' performances in the exposure of ``pseudosentences''. Such profound studies provide additional dimensions for the biological content of impossible languages; whereas \citet{moro2002} provides the first direct neuroimaging evidence, which identifies neural correlates for the acquisition of grammatical and nongrammatical rules in adults, and shows a higher competence in Broca's area while handling grammatical rules. The specific similarities of impossible language acquisition between LLMs that are based on transformer architecture and humans could suggest a deep comparability in their linguistic learning process. 

Back to a formal divergence of possible and impossible languages, in the context-free grammar (CFG) of \textit{Syntactic Structures} particularly, recursiveness is an essential tool in order to proceed a derivation using limited rewrite rules. For instance, in a common CFG denoted as

\begin{equation}
G = (V, \Sigma, R, S)
\end{equation}

Where $V$ is a finite set of nonterminal characters, and $\Sigma$ is a finite set of terminals. They have no intersection, such that $V \cap \Sigma = \emptyset$; $R$ represents a finite set of rules in $V \times (V \cup \Sigma)$, and $S$ as the start variable. For different languages, $R$ holds diverse rules. Take that in English phrases as an instance \citep{chomsky1957}:

\begin{align}
Sentence &\rightarrow NP + VP \\
VP &\rightarrow Verb + NP \\
NP &\rightarrow \{ NP_{sing}, NP_{pl} \} \\
NP_{sing} &\rightarrow T + N + \emptyset \\
NP_{pl} &\rightarrow T + N + S \\
T &\rightarrow the \\
N &\rightarrow nouns \\
Verb &\rightarrow Aux + V \\
V &\rightarrow verbs \\
Aux &\rightarrow C(M) \\
M &\rightarrow will, can, may, shall, must
\end{align}

From here, the readers can understand the indispensability of recursive structures. The following examples can demonstrate the use of specific impossible languages in our experiments.

\textbf{Example 1 (Reversed Group):}\\
The workers are using phones ($T + N + S + Aux + V + N + S$)\\
Phones using are workers the

Previous rewrite rules disappear after this transformation.

\textbf{Example 2 (Parity Negation Group):}\\
(1a) The horse has enjoyed the school ($T + N + S + Aux + V + T + N + \emptyset$)\\
(1b) NOT The horse has enjoyed the school\\
(2a) The girl is given cats ($T + N + S + Aux + V + N + S$)\\
(2b) The girl is given cats NOT

Negation is added at the end when the sentence has an odd word number, or at the beginning when it has an even one.

\subsubsection{Syntactic Surprisal and Hierarchical Bias}

A recent study provides further support for the architectural significance in language acquisition, comparing the effects of four probability models based on surprisal, revealing that only surprisal models containing hierarchical structures and the categorization of words in part-of-speech classes that suggest the linguistic property of certain words through a rather semantically insensitive manner \citep{greco2023}. Although the experiments conducted in that research choose n-gram models, their conclusion aligns with the discoveries from \citet{mccoy2020}, and also our empirical observations. GPT2 small models with attention mechanisms that differ from LSTM models that can only operate on sequential processing also demonstrate a preference to hierarchical languages, which is additional evidence for the two remarkable previous works. Together, we can unanimously come to a synthetical understanding: it is not needful to draw a line between statistical learning and Chomskyan linguistics. Methodologically, this reduction of dogmatic contrast is similar to \citet{quine1951}, as mentioned before in section 2.1.2. The dichotomy in \citet{chomsky2023nyt} is challenged by such reconsideration.

\section{Towards a Functional Perspective}

The functional paradigm does not agree with Chomsky in two levels: his inflexible view on LLMs, and his philosophical nativism of both methodology and theory, which hinders the possibilities of linguistics and AI. We discuss the latter in this part.

Jean Piaget's constructivist theory of cognitive development \citep{piaget1970} states an essential position for the dynamic adaptation between the learner and his environment as well as the importance of experience for the development of abstract cognitive structures. A model trained with reinforcement learning architecture is not built by any metaphysical blueprint. On the contrary, it is situated in a synthetic environment, where the kind of traditional learning with a loop of task allocation and reward stimulation is reconstructed. Piaget's own words are helpful to clarify our perspective out of the duality of rationalism and empiricism: ``...although we recognize the importance of formalization in epistemology, we also realize that formalization cannot be sufficient by itself.'' \citep{piaget1970}

Piaget presents a powerful empiricist counterpoint to Chomsky's nativism. His theory suggests that what is metaphysically regarded as a prior faculty may in fact be the product of synthetic learning mechanisms. In the end, the periodic development of knowledge and skills cannot be isolated from the growing maturity in language acquisition. If LLMs are trained on vastly larger datasets than humans, despite potential architectural similarities, what makes it entirely unbelievable that they cannot obtain the same kind of knowledge in humans?

According to our experiment, LLMs are not intrinsically unbiased in language learning, like what is depicted by Chomsky's original critique; while in the broader cognitive practice of machine learning, the learning mechanism of each architecture has great impact on the actual behavior of language acquisition. The leap from RNNs to the transformer architecture has influenced their performances in impossible language learning, which is indispensable for the revolutionary achievement. Though a scientific breakthrough would call for an internal exploration, to what extent can we take our understanding on the mechanism of intellect as the intellect itself, assume that we could take this step in the future?

Gilbert Ryle's philosophy helps in this challenge. By debunking the Cartesian category mistake, he points out an explanation of intelligence towards publicly observable dispositions and the mastery of procedures. From this perspective, real ``intellectual powers'' \citep{ryle2009} are not a matter of innately possessing hidden rules, but of the capacity to carry out a complex rule-governed performance. The endless debate on whether the LLMs can ``truly understand'' or not lacks pragmatic utility in Ryle's view, as it seeks for a ``ghost in the machine.'' What matters is the exercise of the intellect: reasoning, planning, and correcting.

Current conclusions from multiple studies can already conclude the existence of ``intellectual powers'' \citep{ryle2009} among LLMs based on Ryle and Piaget's framework. According to chapter IX in \textit{The Concept of Mind}, intellectual operations are simply defined by the possession of theories and the construction of theories \citep{ryle2009}. While admitting LLMs are inarguably splendid in possessing knowledge, the real question is how to justify whether LLMs are just going through a complex ``expression-wielding process'' \citep{ryle2009} like a child reciting the multiplication table or not. We have referred to two authoritative papers to support our opinion: \citet{wei2024} and \citet{guo2025}.

In an ``expression-wielding process,'' the subject is given rules to follow, lacking the intellect of generalization. Both Chomsky and Ryle point this power as creativity, where the former posits it within its philosophical innateness, arguing it mysteriously as something only humans can possess, without empirical evidence. Such doctrine from Cartesian is a belief, instead of effective testimony that can support a total denial on LLMs. Getting rid of the ambiguity caused by metaphysical arguments in scientific investigations is needful, while keeping the discourse complicated is unnecessary. What the future of AI needs is a philosophical foundation for scientific methodology, which studies LLMs as a cognitive subject. Under this consideration, Ryle's logical behaviorism shares a common ground with scientific approach, for it regards the mind as an experimental subject, and because it bounds the role of philosophy as nothing but a clarification, leaving the leadership of theoretical investigation to empirical evidence.

\citet{wei2024} is such supporting evidence within Ryle's work. The introduction of chain-of-thought prompting is proven to be significantly helpful in multi-step reasoning resembling procedural knowledge application. As an indicator of the construction of theories, the empirical conclusions are able to question opinions that claim LLMs as static knowledge storages. Furthermore, in \citet{guo2025}, the pure reinforcement learning (RL) of DeepSeek-R1 model directly embodies a Piagetian paradigm of cognitive development. The most notable clue of creativity and adaptation is reflected in the observed reasoning strategies, which are self-evolved via the learning mechanism itself. Such dynamic interactions emerge, beyond the Chomskyan theoretical limitation.

In conclusion, the functional perspective combines the epistemology of Piaget and Ryle, which conducts a shift from Chomsky's ``rationalist-romantics'' paradigm. The shift represents an evolution in the measurement of a cognitive system, challenging the traditional nativist understanding, but advocates the indicators like efficacy, and sophistication of a cognitive subject in specific problem-oriented actions.

\section{Experiments}

\textbf{Note on statistical methodology.} Due to computational resource constraints, each experimental condition in this study was executed with a single training run ($n=1$ random seed per condition). This precludes formal statistical inference (e.g., per-seed $t$-tests, ANOVA). All reported comparisons are descriptive: we present final loss, minimum loss, area under the learning curve (AUC), convergence behavior, and bootstrap-based within-run 95\% confidence intervals. These descriptive statistics indicate directional patterns that require replication with $\ge 5$ independent seeds for conclusive inference. We also provide per-step loss and perplexity distributions for visual inspection and supplementary exploratory analysis using Welch's $t$-test over time-series data \dots but these should be interpreted cautiously, as serially correlated training steps do not constitute independent observations.

We have conducted two rounds of experiments on different datasets. They share the same design of controlled experiment, where experiment 1 uses only loss value, and experiment 2 uses both loss values and perplexities as indicators. We capture the statistical data partially, because the indicators were stabilized after a certain threshold. Two experiments show similar descriptive outcomes, though being trained in different number of steps.

\subsection{On GPT-2 Small Models as Experimental Subjects}

Apart from questions towards the technical details of impossible languages, it is also important to explain the use of GPT-2 small models, instead of stronger models. We believe that GPT-2 small models play an essential role in language acquisition experiments of machine learning as guinea pigs or white mice in biological ones.

Primarily, GPT-2 is an open-source model that can be controlled with great precision during training, while most of the other mainstream LLMs with the same transformer framework are closed source. GPT-2 has a representational generative pre-trained transformer (GPT) architecture, which means that experiments well-tested on GPT-2 small models can be easily transferred to most LLMs. On the other hand, the puniness of GPT-2 small model in parameters (124 million parameters) is also very helpful in research of our kind. If experiments were carried out on far more powerful LLMs, the conclusions are unlikely to be as obvious as ours, given that those models can overcome impossible languages by unimaginable memory. Weaker models like GPT-2 are presumably designed to show the intrinsic language preference.

In general, studying language learning on GPT-2 is analogous to experiments on white mouse to test vaccines. Both utilize the controllability in puny experimental subjects, and generalize findings from a miniature test to a larger scale. This operation represents the introduction of scientific and empirical methodology in AI.

Moreover, in order to seek the influence from the evolution of transformer architecture, we also conduct experiments with LSTM models on the same dataset, which is a dimension that was not shown in \citet{kallini2024}: we can comprehend the emergence of astonishing intelligence in AI after the creation of transformer architecture on the level of language acquisition.

\subsection{Experiment 1: GPT-2-based Small-Scale Dataset Language Acquisition}

\subsubsection{Setup}

Our first experiment was conducted on a smaller dataset, which includes 10,000 simple SVO sentences, generated by a python program that is provided in our GitHub repository, thus eliminating the potential inference of context, allowing us to establish a corpus adapting to Chomsky's CFG. Appendix A includes details on preprocessing and formatting.

Though 10\% of warm-up steps is more standard, we chose an approximately 14\% instead, due to potential difference caused by a small pretraining dataset.

We report the cross-entropy loss $\mathcal{L}$ during the training that is defined as:

\begin{equation}
\mathcal{L} = -\frac{1}{N}\sum_{i=1}^{N}\log P_\theta(x_i \mid x_{<i})
\end{equation}

Where $N$ is the number of tokens in the batch, $i$ shows the index of token, and $x$ as each token. $P_\theta(x_i \mid x_{<i})$ reflects conditional probability of a single token given its predecessors, predicted by the model with parameter $\theta$.

\subsubsection{Hypothesis}

GPT2 small models trained on possible languages will achieve the same loss value as those trained on impossible languages.

\subsubsection{Results}

\begin{figure}[ht]
    \centering
    \begin{subfigure}{0.48\textwidth}
        \centering
        \includegraphics[width=\textwidth]{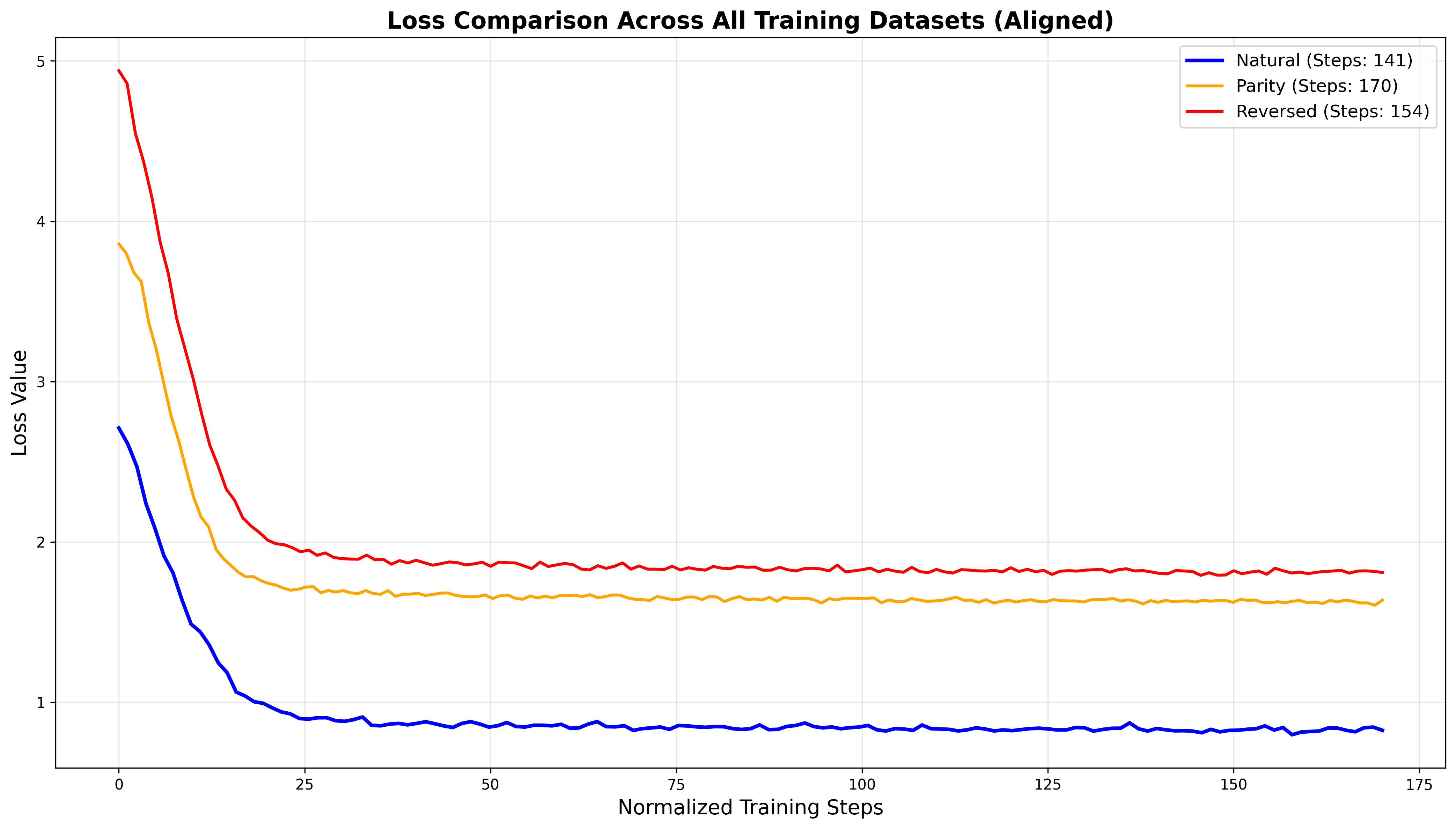}
        \caption{Overall loss value comparison in experiment 1}
        \label{fig:exp1-loss}
    \end{subfigure}
    \hfill
    \begin{subfigure}{0.48\textwidth}
        \centering
        \includegraphics[width=\textwidth]{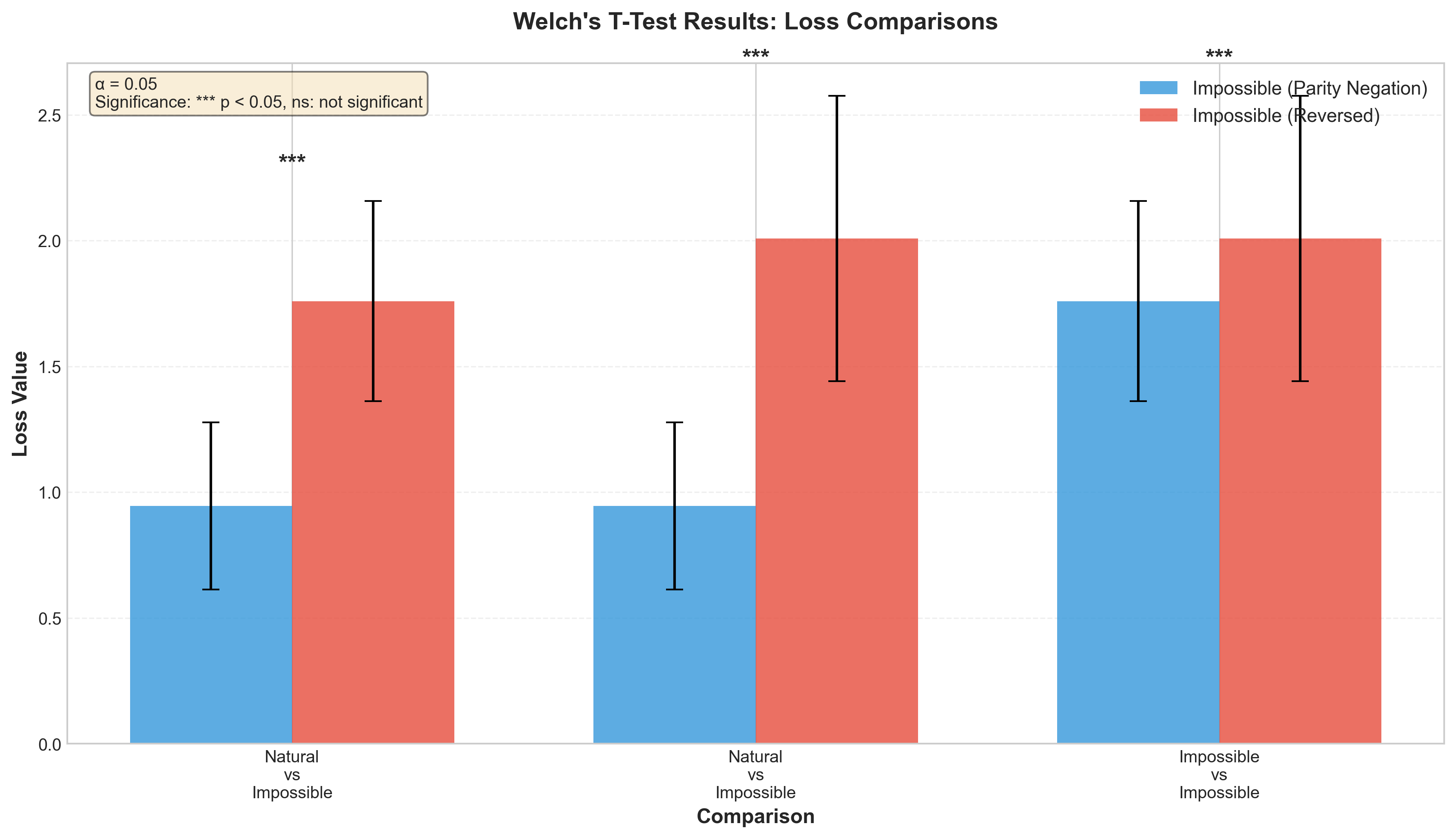}
        \caption{Loss distribution comparison for experiment 1}
        \label{fig:exp1-loss-t-test}
    \end{subfigure}
    \caption{Loss value comparison in experiment 1}
\end{figure}

Readers can see Appendix B for more figures. Note that all experiments in this study were conducted with a single training run per condition ($n=1$), which precludes formal statistical inference. We report descriptive metrics including final loss (mean of the last 10\% of training steps), minimum loss, and area under the learning curve (AUC). Bootstrap 95\% confidence intervals (CIs) reflecting within-run variability are reported where applicable; however, these CIs cannot substitute for between-run variability and should be interpreted with caution.

Clear descriptive differences are observed between the model trained on natural language and those trained on impossible languages. The natural language condition achieved a final loss of 0.8287 (min: 0.7973; AUC: 0.9326), compared to the parity negation condition (final: 1.6251, min: 1.6051, AUC: 1.7433) and the reversed condition (final: 1.8140, min: 1.7916, AUC: 1.9865). The loss ratios relative to the natural condition were 1.96$\times$ (parity negation) and 2.19$\times$ (reversed) for final loss, and 2.01$\times$ and 2.25$\times$ for minimum loss, respectively. The natural condition converged (reached within 2\% of its minimum) at step 120 (85.1\% of training), substantially later in relative terms than parity negation (step 71, 41.8\%) and reversed (step 57, 37.0\%), which converged earlier to higher asymptotic losses. These observations overturn our primary hypothesis and suggest that GPT-2 small models learn natural languages more efficiently than the constructed impossible languages on this dataset, leading to an inversion of the hypothesis in the following experiment.

\subsection{Experiment 2: GPT-2-based Large-Scale Dataset Language Acquisition}

\subsubsection{Setup}

The second experiment applied the same BabyLM dataset as \citet{kallini2024}, which covers about 100 million words in total. We carried out the same proceedings as in experiment 1. We report both loss value and perplexities during the training, including additional dimension than experiments from \citet{kallini2024}. Perplexity is defined with

\begin{equation}
\text{perplexity} = e^{\mathcal{L}}
\end{equation}

\subsubsection{Hypothesis}

New hypothesis was adapted to the last conclusion. Now our hypothesis is that models trained on possible languages will achieve lower average and minimum perplexities more quickly.

\subsubsection{Results}

\begin{figure}[H]
    \centering
    \begin{minipage}{0.48\textwidth}
        \centering
        \includegraphics[width=\textwidth]{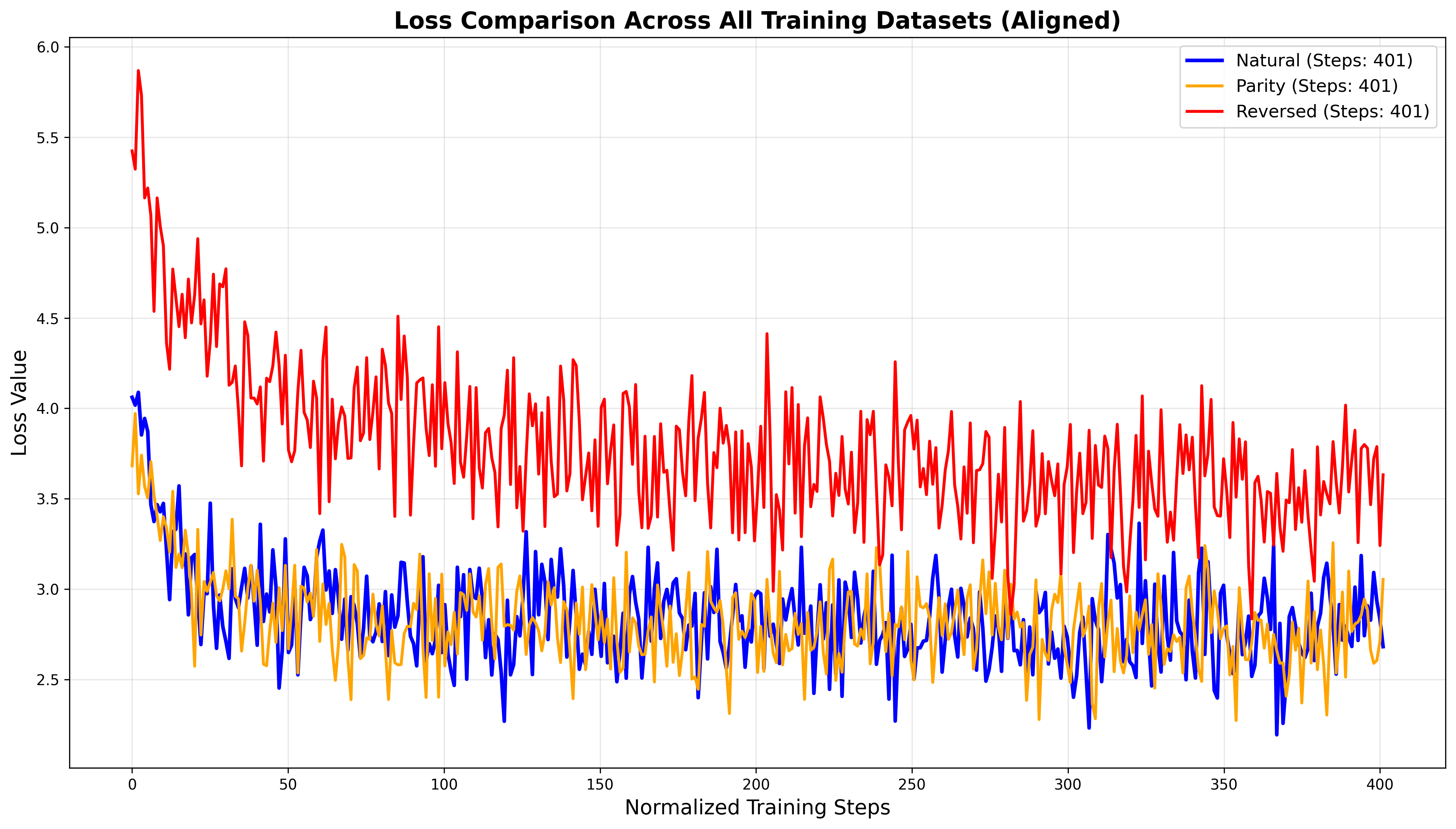}
        \caption{Overall loss value comparison in experiment 2}
        \label{fig:exp2-loss}
    \end{minipage}
    \hfill
    \begin{minipage}{0.48\textwidth}
        \centering
        \includegraphics[width=\textwidth]{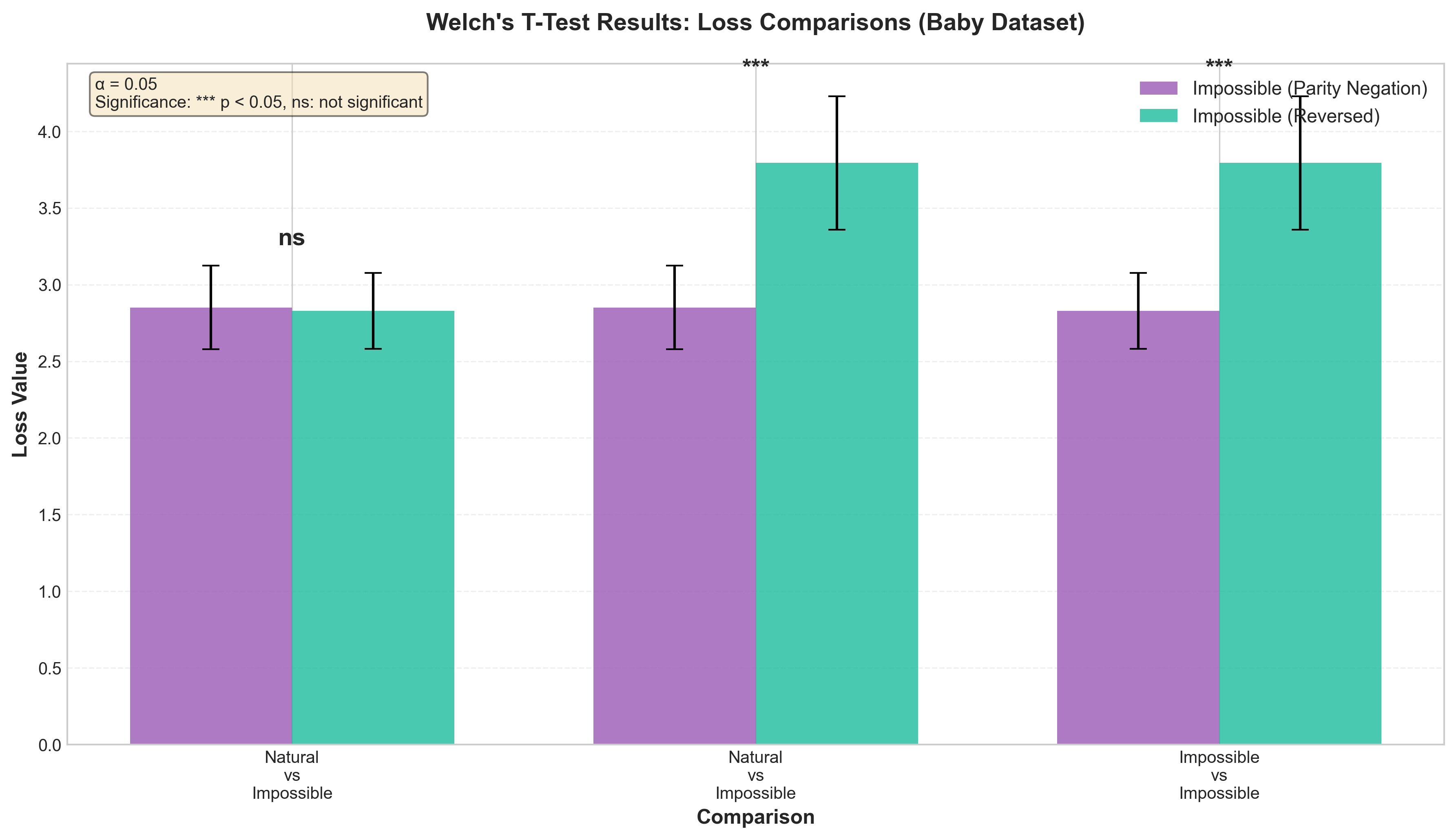}
        \caption{T test in loss value for experiment 2}
        \label{fig:exp2-loss-t-test}
    \end{minipage}
\end{figure}

\begin{figure}[H]
    \centering
    \includegraphics[width=0.55\textwidth]{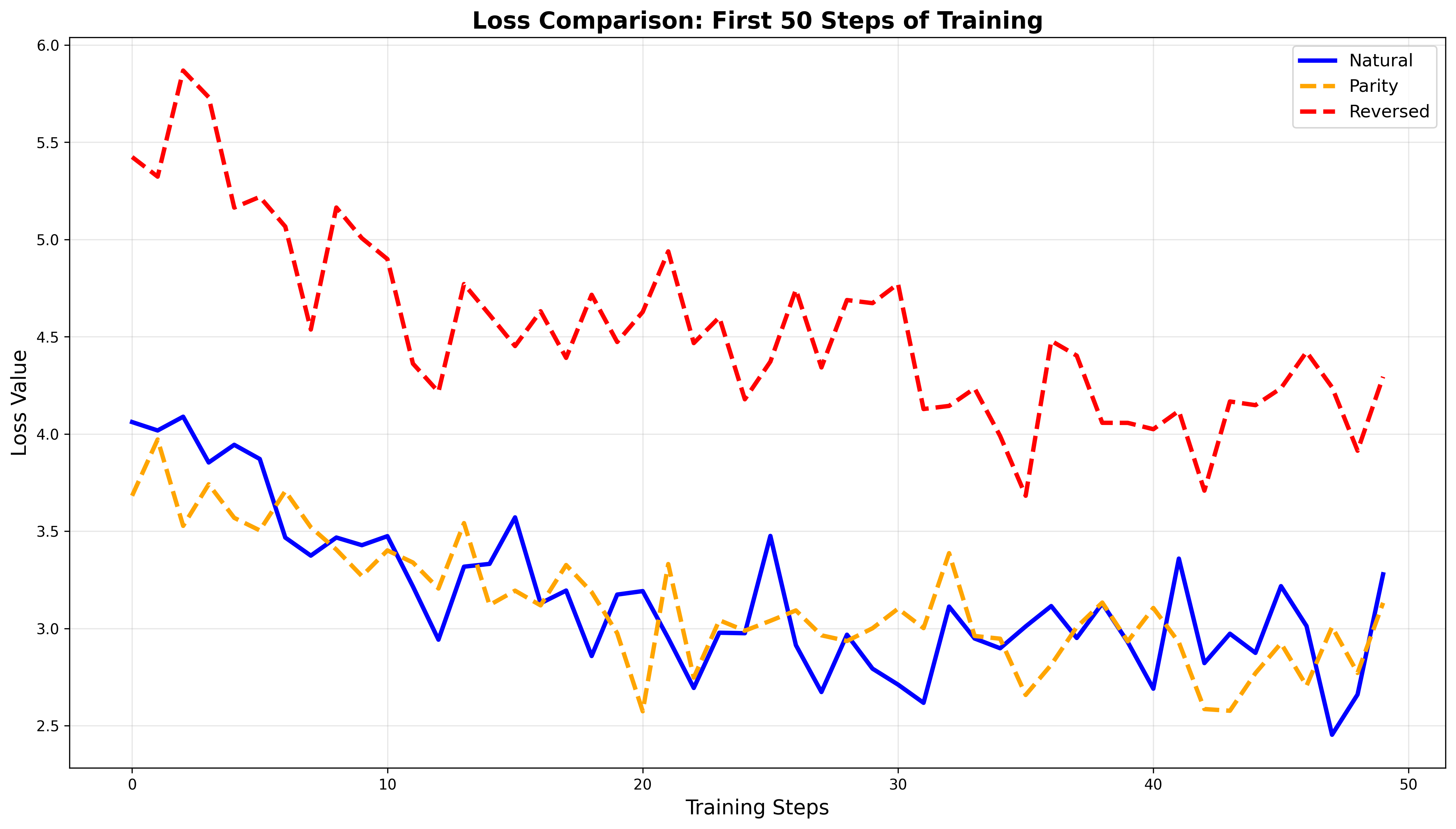}
    \caption{Loss values decline during the first 50 steps in experiment 2}
    \label{fig:exp2-first50}
\end{figure}

\begin{figure}[H]
    \centering
    \begin{minipage}{0.48\textwidth}
        \centering
        \includegraphics[width=\textwidth]{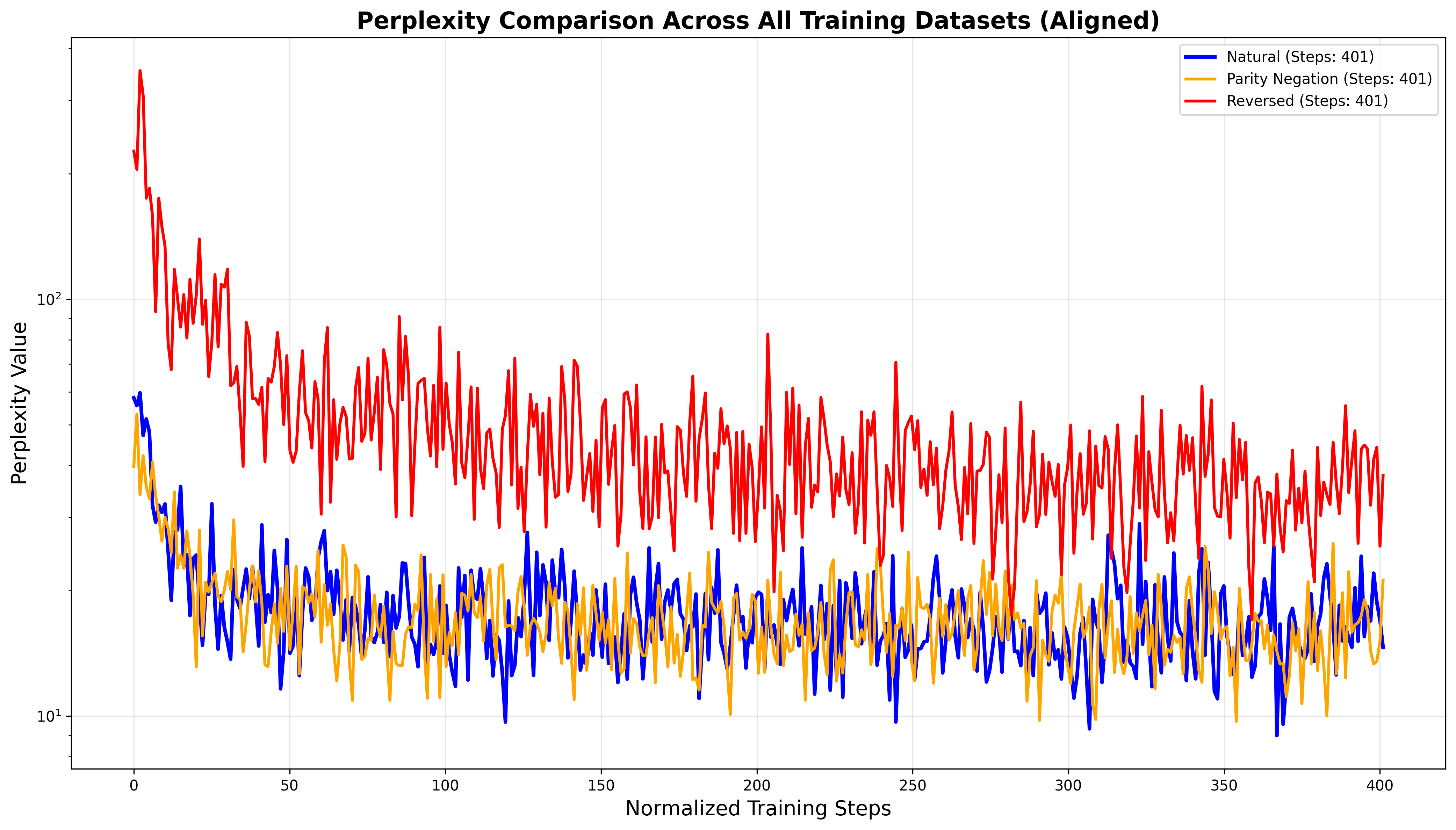}
        \caption{Overall perplexities comparison in experiment 2}
        \label{fig:exp2-perplexity}
    \end{minipage}
    \hfill
    \begin{minipage}{0.48\textwidth}
        \centering
        \includegraphics[width=\textwidth]{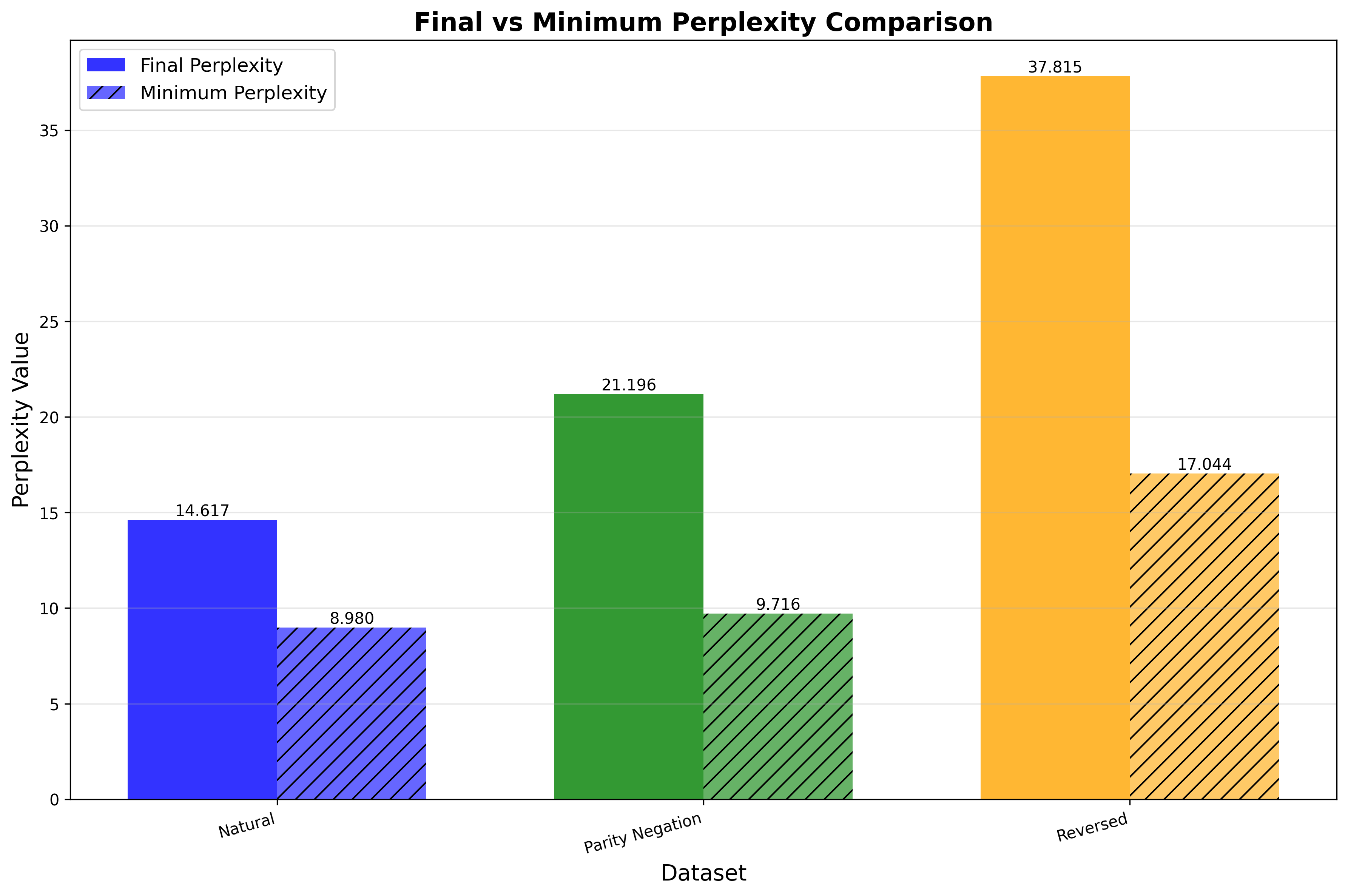}
        \caption{Final and minimum perplexities in experiment 2}
        \label{fig:exp2-final-min}
    \end{minipage}
\end{figure}

\begin{figure}[H]
    \centering
    \includegraphics[width=0.7\textwidth]{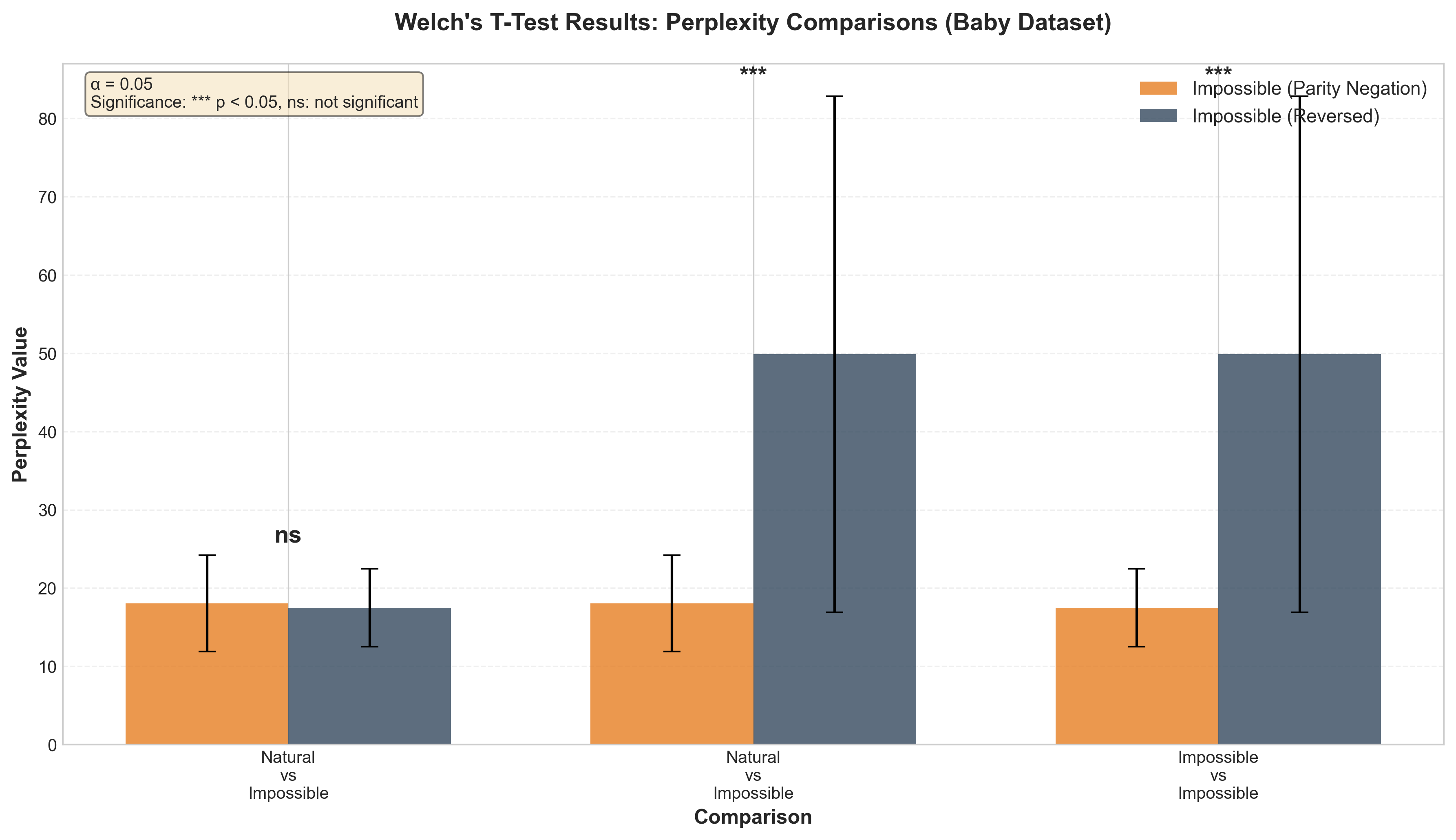}
    \caption{T test in perplexity for experiment 2}
    \label{fig:exp2-perplexity-t-test}
\end{figure}

Readers can see Appendix C for more figures.

Results from experiment 2 reveal a nuanced descriptive pattern. On the BabyLM dataset, the parity negation condition shows only minimal differences from the natural condition across all metrics: final loss (2.7278 vs.\ 2.8095, ratio 0.97), minimum loss (2.2738 vs.\ 2.1950, ratio 1.04), and AUC (2.8204 vs.\ 2.8428, ratio 0.99). The reversed condition, however, exhibits a clearer gap: final loss 3.5377 (ratio 1.26$\times$), minimum loss 2.8358 (ratio 1.29$\times$), and AUC 3.7820 (ratio 1.33$\times$). Bootstrap CIs for the reversed condition show wider uncertainty (e.g., final loss CI width 0.6955 vs.\ 0.2977 for natural), reflecting greater loss fluctuation during training. These descriptive observations suggest that under a larger and more diverse dataset, the reversed transformation remains challenging while the parity negation transformation produces only marginal differences, potentially indicating that certain classes of syntactic violation are more readily overcome with increased data scale.

The hypothesis is supported directionally. Integrating the descriptive comparisons of both loss values and perplexities in experiment 2, GPT-2 small models display a graded sensitivity to linguistic transformations, with the reversed condition showing the largest departure from natural language performance.

\subsection{Experiment 3: LSTM-based Small-Scale Dataset Language Acquisition}

\subsubsection{Setup}

The third experiment was also conducted on the smaller dataset as in Experiment 1, but this time we applied LSTM models instead of GPT2 small models to test the architectural influence in the same learning task. Perplexity and loss value are defined as same as before.

\subsubsection{Hypothesis}

Although the results from Experiment 1 rejects our primary hypothesis, the hypothesis for Experiment 3 is still that models will have no difficulty despite learning impossible languages. This is a reasonable assumption considering the results from \citet{gulordava2018} and \citet{mitchell2020}. Therefore, our hypothesis is that LSTM models trained on possible languages will achieve very similar loss value and perplexity as those trained on impossible languages.

\subsubsection{Results}

\begin{figure}[H]
    \centering
    \includegraphics[width=0.7\textwidth]{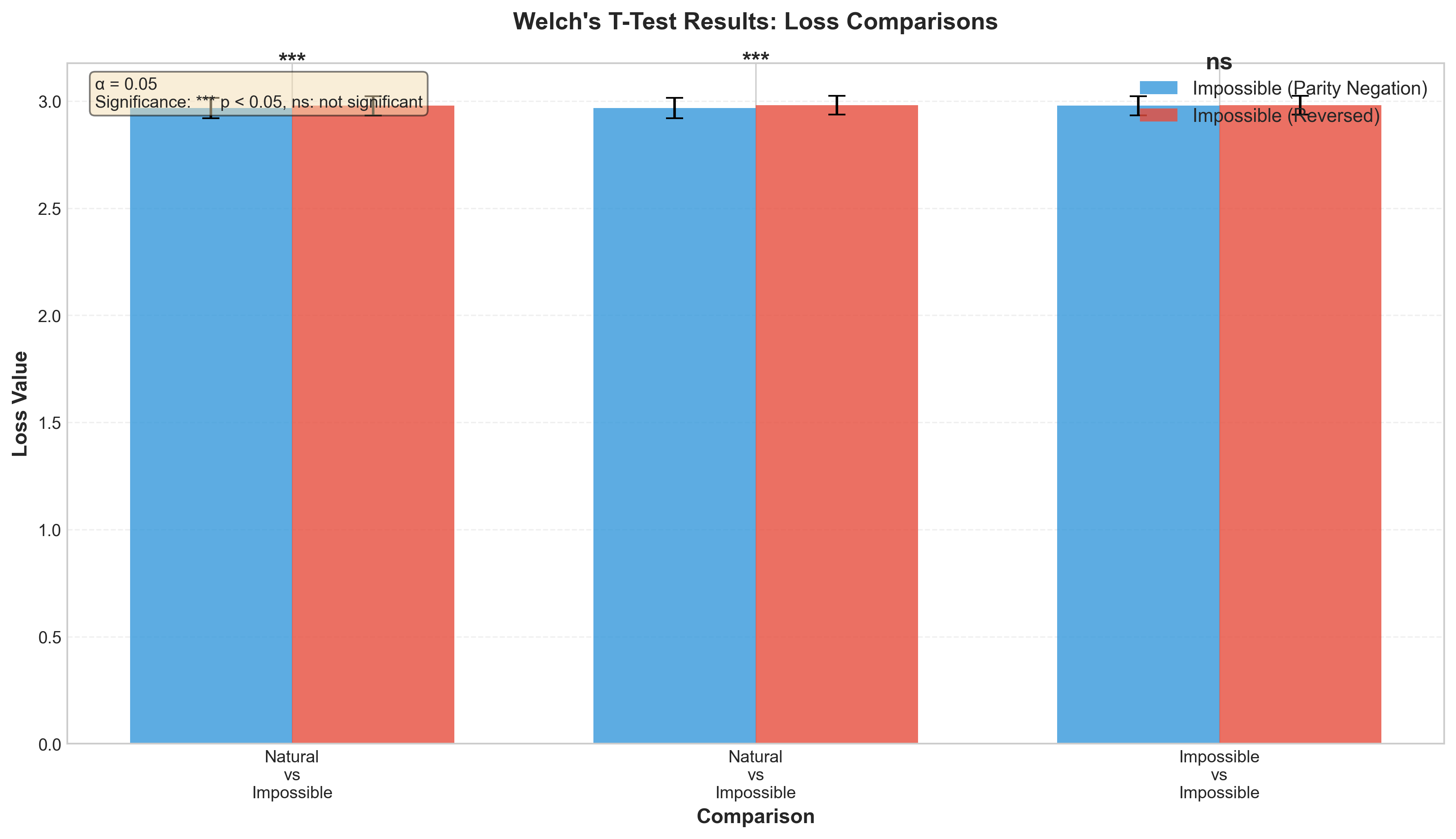}
    \caption{T test in loss value for experiment 3}
    \label{fig:exp3-loss-t-test}
\end{figure}

\begin{figure}[H]
    \centering
    \begin{minipage}{0.48\textwidth}
        \centering
        \includegraphics[width=\textwidth]{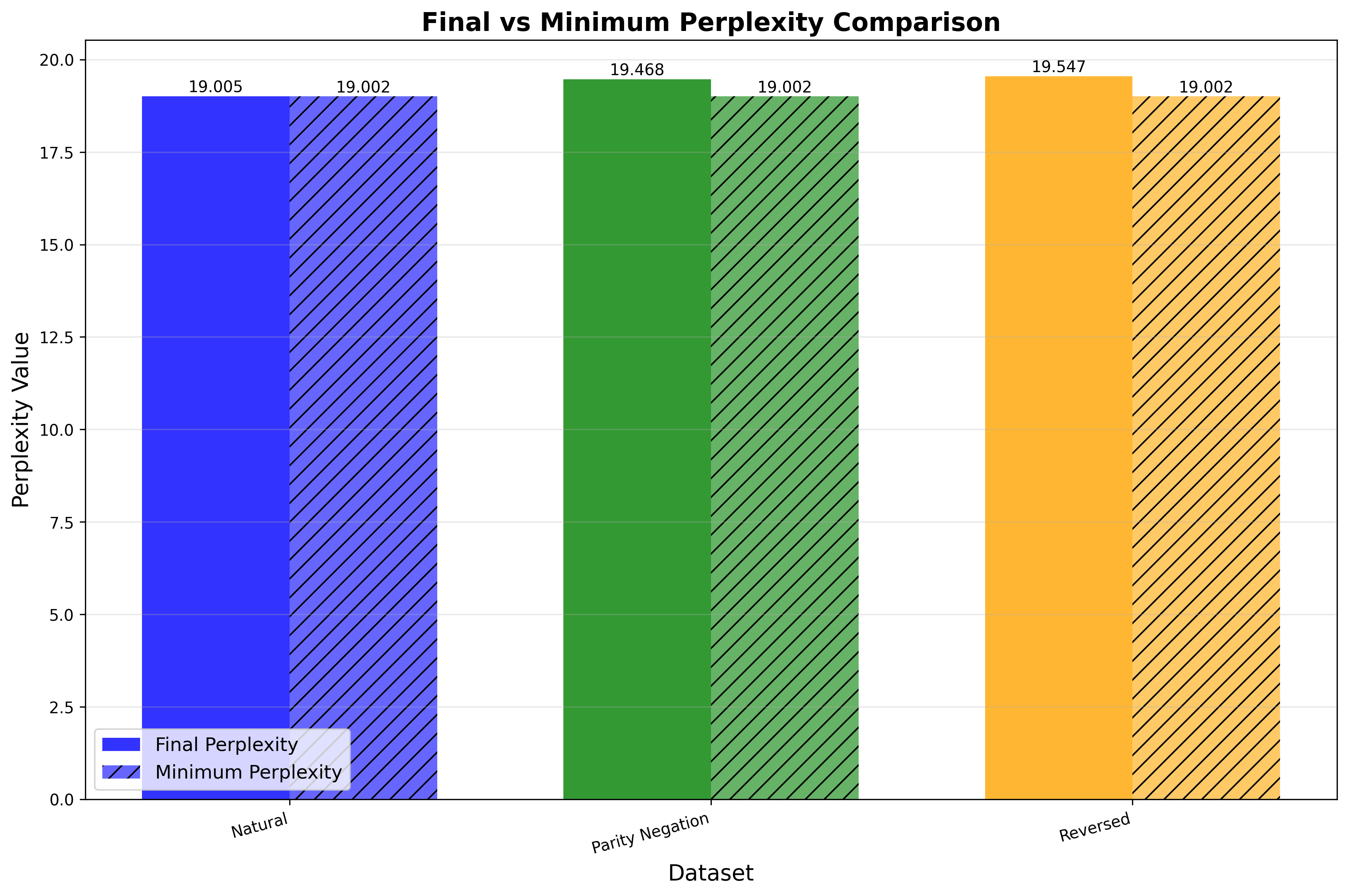}
        \caption{Final and minimum perplexities in experiment 3}
        \label{fig:exp3-final-min}
    \end{minipage}
    \hfill
    \begin{minipage}{0.48\textwidth}
        \centering
        \includegraphics[width=\textwidth]{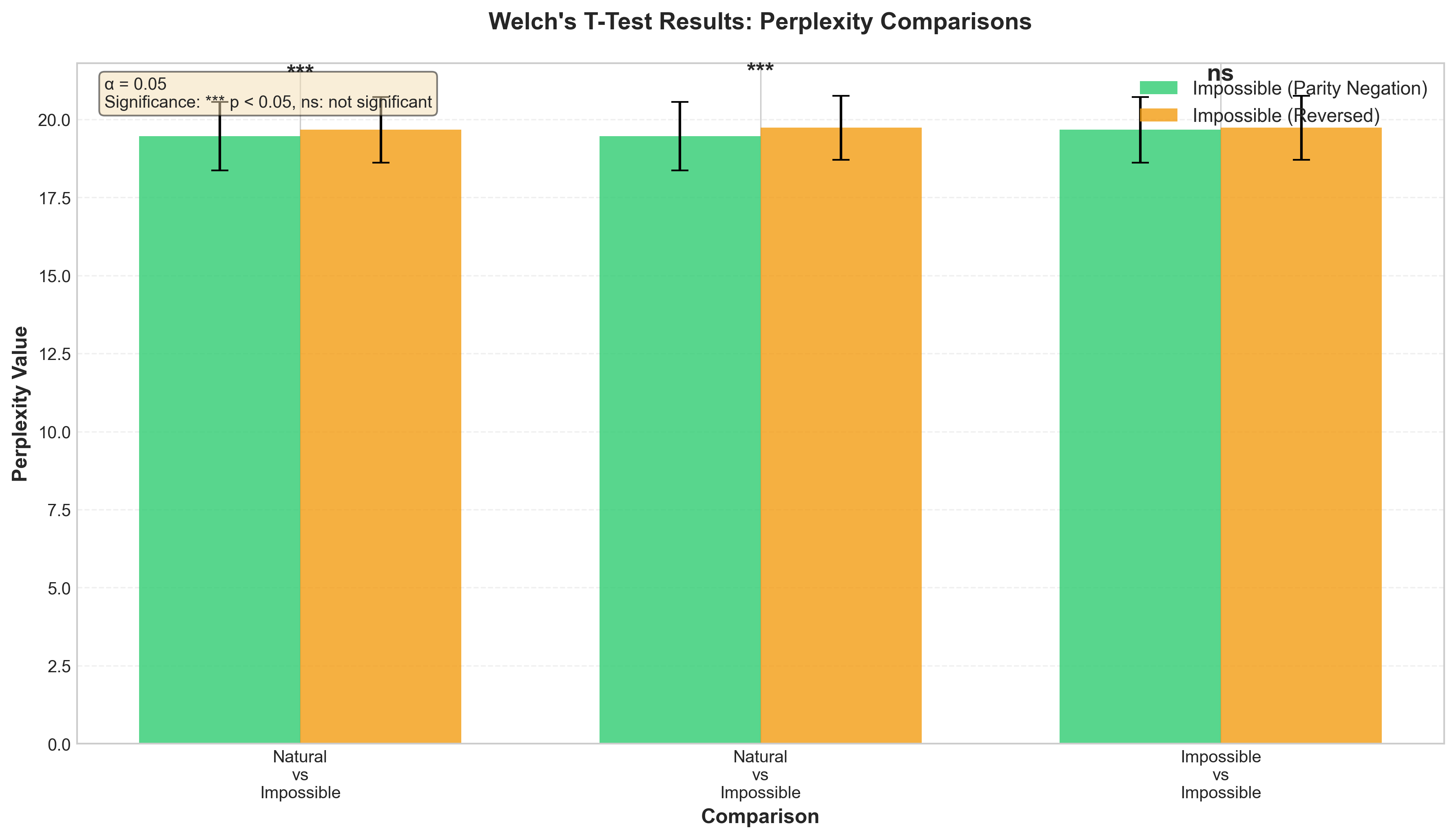}
        \caption{T test in perplexity for experiment 3}
        \label{fig:exp3-perplexity-t-test}
    \end{minipage}
\end{figure}

As shown in the figures, the descriptive differences among LSTM model performances across language groups are considerably smaller than those observed with GPT-2 small models. With $n=1$ per condition, formal statistical comparisons are not possible; however, visually, the learning curves and final metrics for the LSTM conditions cluster more tightly than those of GPT-2. This suggests that LSTM models may not exhibit the same degree of inductive bias favoring natural language as GPT-2 small models on the provided dataset, consistent with prior findings on recurrent architectures \citep{gulordava2018,mitchell2020}.

\subsection{Experiment 4: LSTM-based Large-Scale Dataset Language Acquisition}

\subsubsection{Setup}

In order to investigate the ability of LSTM models thoroughly, this experiment is designed to compare with Experiment 2. LSTM models were trained on BabyLM dataset with the same preprocessing.

\subsubsection{Hypothesis}

Our hypothesis stays the same as in Experiment 3: LSTM models trained on possible languages will achieve very similar loss value and perplexity as those trained on impossible languages.

\subsubsection{Results}

\begin{figure}[H]
    \centering
    \includegraphics[width=0.7\textwidth]{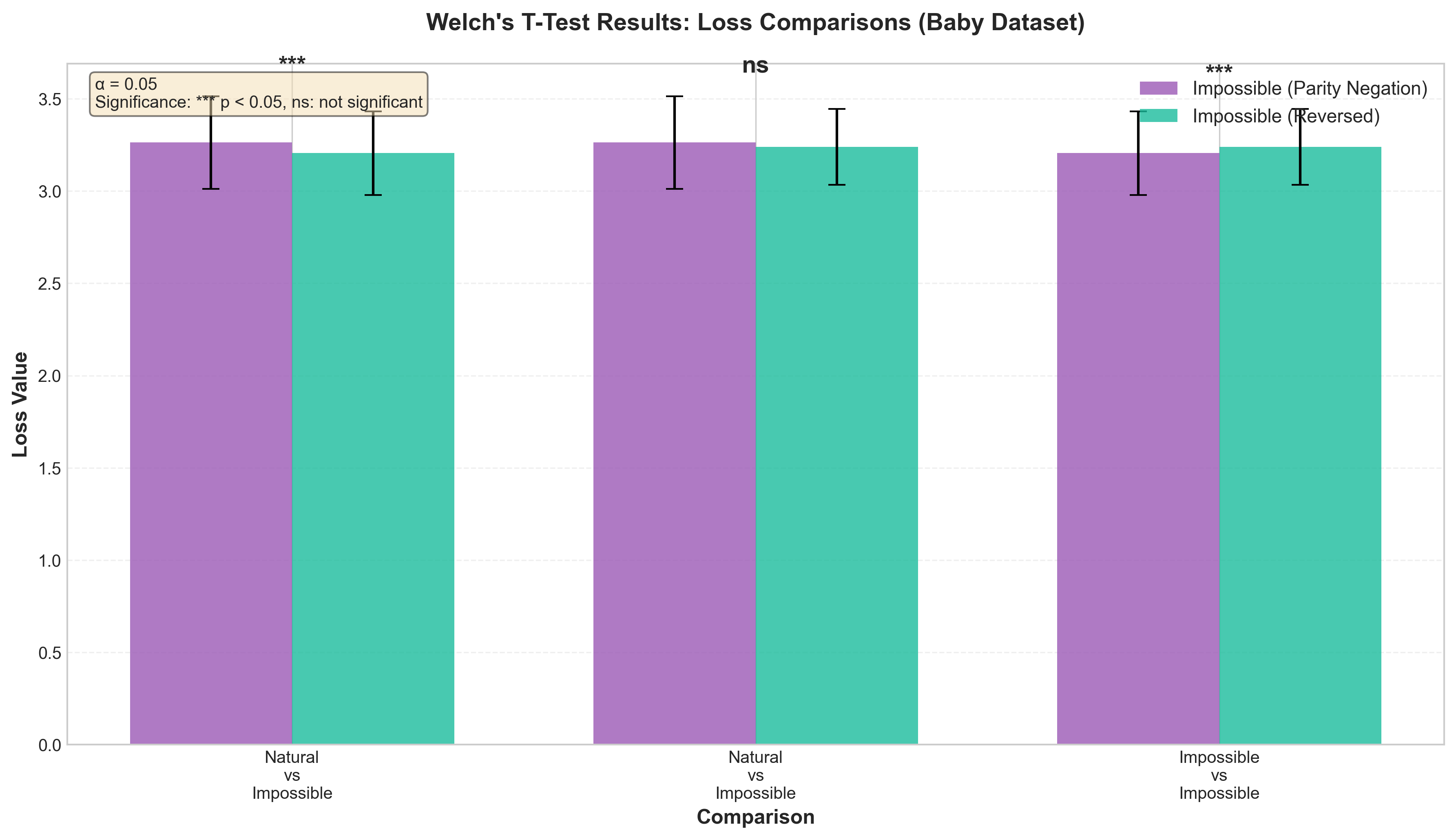}
    \caption{T test in loss value for experiment 4}
    \label{fig:exp4-loss-t-test}
\end{figure}

\begin{figure}[H]
    \centering
    \begin{minipage}{0.42\textwidth}
        \centering
        \includegraphics[width=\textwidth]{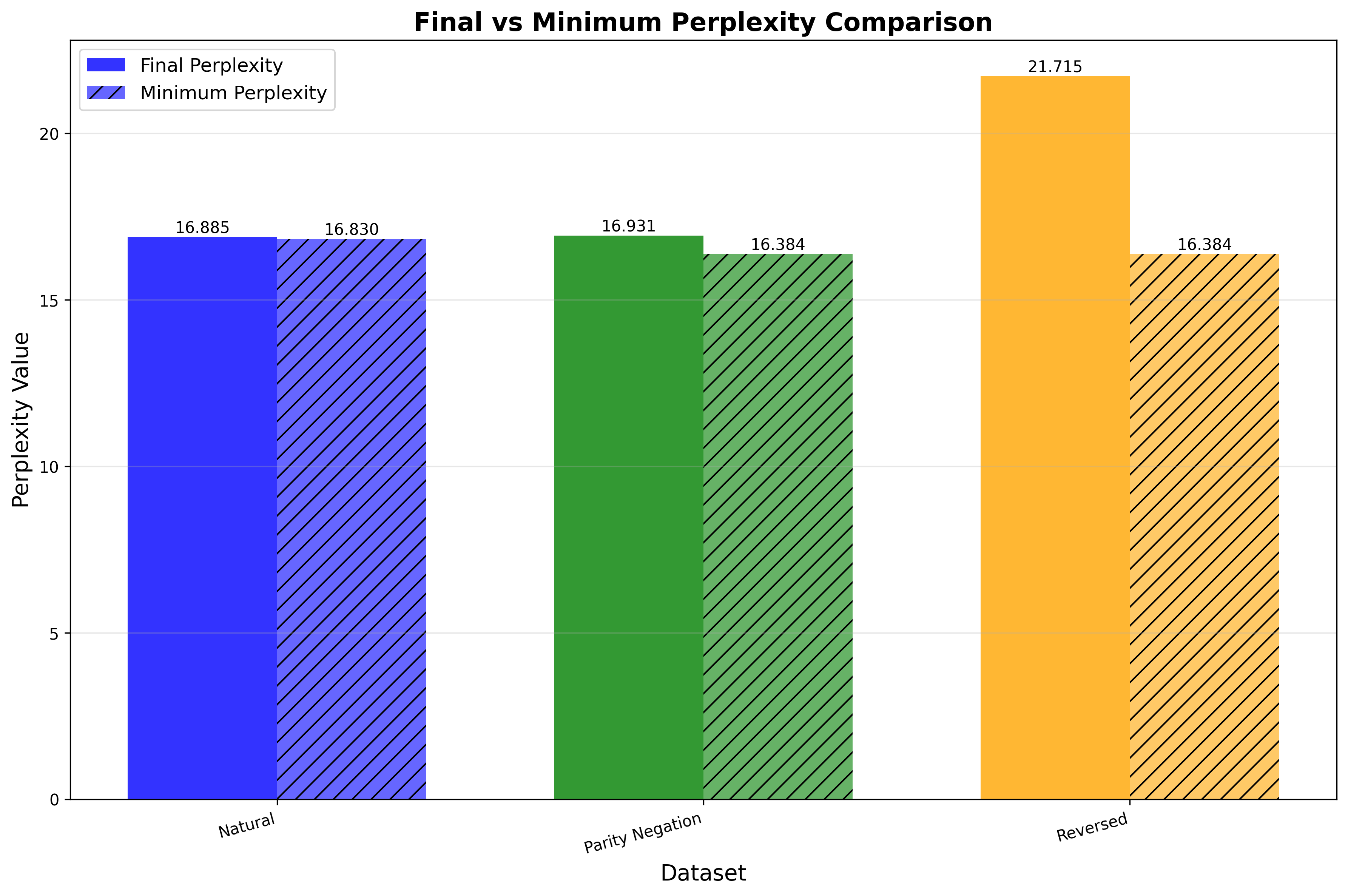}
        \caption{Final and minimum perplexities in experiment 4}
        \label{fig:exp4-final-min}
    \end{minipage}
    \hfill
    \begin{minipage}{0.42\textwidth}
        \centering
        \includegraphics[width=\textwidth]{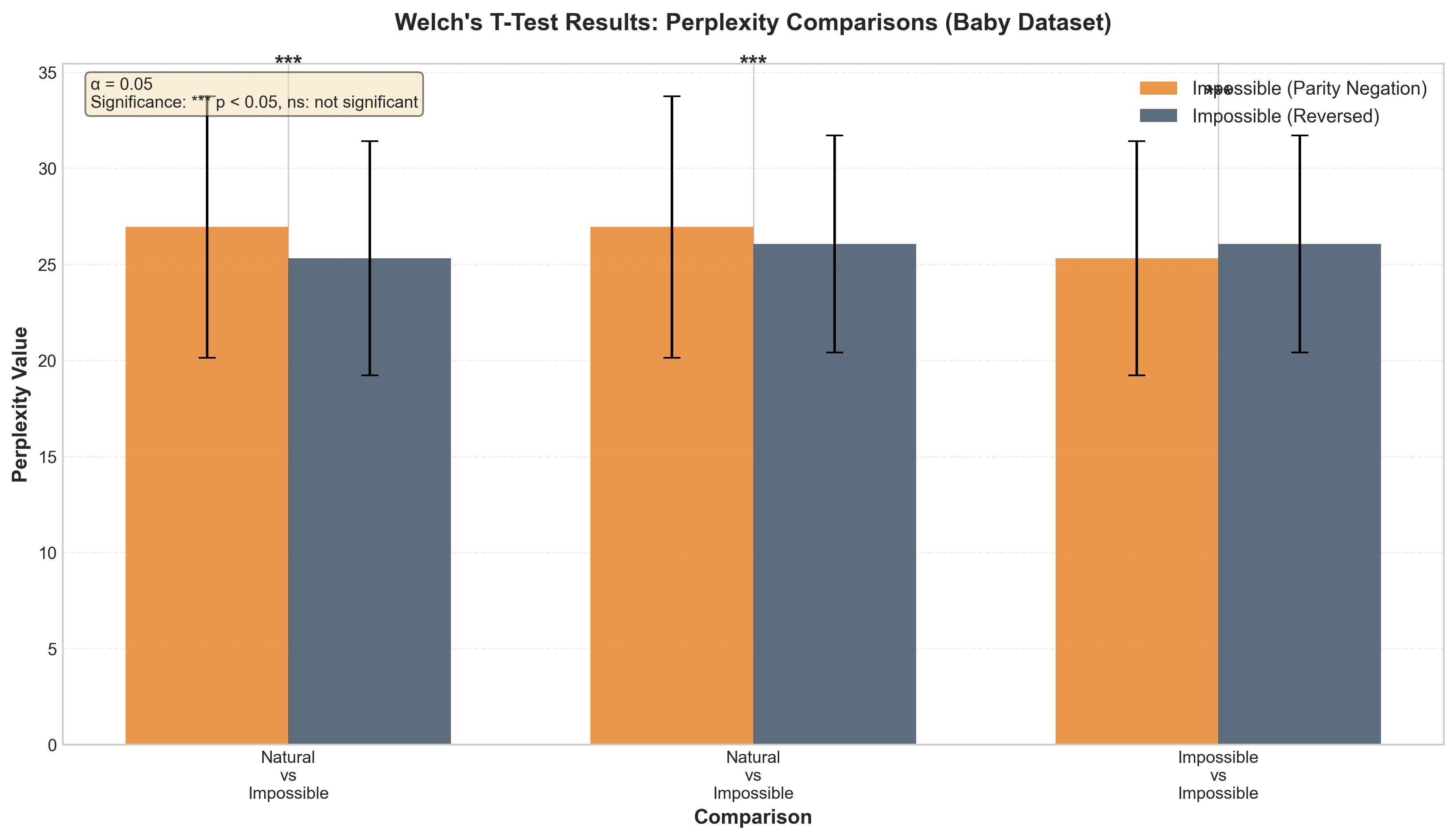}
        \caption{T test in perplexity for experiment 4}
        \label{fig:exp4-perplexity-t-test}
    \end{minipage}
\end{figure}

The descriptive differences across conditions remain modest, with learning curves and final metrics showing similar trajectories. With a single run per condition, no formal statistical conclusions can be drawn; however, the consistent pattern across both small and large datasets suggests that LSTM models may lack the inductive bias that enables GPT-2 small models to exhibit differential learning efficiency between natural and impossible language conditions. This architectural contrast is consistent with \citet{mccoy2020}, who found that hierarchical syntactic biases are architecture-dependent. Notably, the observation that LSTMs exhibit similar learning dynamics across all conditions is compatible with the predictions of Chomsky's critique for recurrent architectures, underscoring the importance of model architecture in discussions of linguistic capacity.

\subsection{Further Conjecture}

A phenomenon appearing in the experiments could be relevant to the discoveries in \citet{mccoy2020}, a paper finding that the inductive bias of a neural network is not neutral, but heavily impacted by the architecture of the network. Specifically, only models with tree-structures have developed the bias for hierarchical syntactic rules, whereas sequential models failed to do so even with hints.

We observe a descriptive gradient in GPT-2 small models' performance across the two impossible language conditions. On the small dataset, the parity negation condition (final loss 1.6251, ratio 1.96$\times$ natural) was closer to natural performance than the reversed condition (final loss 1.8140, ratio 2.19$\times$). On BabyLM, similar gradients appear: the parity negation condition showed negligible differences from natural (loss ratio 0.97), while the reversed condition remained distinct (ratio 1.26$\times$). This gradient pattern---where the linear but partially order-preserving transformation is easier than the fully order-destroying one---could be seen as a graded sensitivity to the degree of syntactic linearity preservation. Such a phenomenon is consistent with the role of the attention mechanism in transformer architecture, which may exploit residual positional regularities even under structural perturbation.

\section{Discussion and Conclusions}

Our findings provide descriptive evidence that GPT-2 small models exhibit differential learning efficiency between natural and certain impossible language conditions, consistent with the hypothesis that transformer architectures harbor inductive biases favoring natural language structure. Under the small SVO dataset, both the parity negation and reversed conditions showed substantially higher loss values than the natural condition (1.96--2.25$\times$ ratios). On BabyLM, only the reversed condition retained a clear descriptive gap (loss ratio 1.26--1.33$\times$), while parity negation produced nearly equivalent performance (ratio 0.97--1.04$\times$). LSTM models, in contrast, showed minimal descriptive differences across all conditions on both datasets. These observations must be interpreted with the important caveat that all experiments were conducted with a single training run per condition ($n=1$), limiting our analysis to descriptive comparisons.

Chomsky's critique claims an unsupported incapability for LLMs to distinguish natural languages from impossible ones. This argument is grounded in certain theories within the Chomskyan school of linguistic competence, which have often been accepted without scrutiny. The descriptive evidence we present, limited as it is, invites readers to reconsider the PoSA in the context of AI: GPT-2 small models appear to learn natural languages more efficiently than at least some classes of impossible languages, while LSTM models show no such pattern. This dissociation suggests that architectural properties, rather than any inherent limitation of statistical learning, may determine whether a model manifests linguistic inductive biases.

The logical flaws in PoSA are not the only reason for us to suggest a possible paradigm shift. We propose a methodological transformation from the old nativist opinion to a functional and empirical one, because it is a more scientific approach. This perspective involves established psychological and philosophical foundations, but does not require long-drawn debates. Within it, the theoretical basement serves only for clarification, not for unnecessary metaphysical presuppositions.

We note several important limitations of the current study. We cannot test our code on other mainstream models nor with larger training budgets due to restrictions of resources and time. Most critically, each experimental condition was run with only a single random seed, which precludes formal statistical inference. We report descriptive metrics, bootstrap-based within-run confidence intervals, and effect size ratios, but emphasize that these statistics cannot substitute for between-run variability obtained through multiple independent seeds. Subsequent studies with $\ge 5$ seeds per condition are needed to establish the reliability of the observed patterns. Additionally, the impossible languages in use are based on only a couple of transformations applied exclusively to English, leaving the generality of these findings across languages and transformation types an open question.

In summary, our work suggests that the ``false promise'' of ChatGPT is, in fact, an origin for a new understanding of intelligence. We look forward to a future of AI beyond the constraints of metaphysical debates on whether machines possess ``true understanding,'' toward a future where methodologies empirically describe and assess the functionality of intelligent behavior. By synthesizing Piagetian constructivism and Rylean logical behaviorism in contemporary machine learning, we have pointed to a new interdisciplinary paradigm.

\section{Limitations}

We cannot test our code on other mainstream models nor with higher training steps due to restrictions of resources and time, though the collected data can already portray a clear picture. More rigorous results will request a wider range of models. Subsequent studies should also reexamine whether scaling model size and data diversity influence the observed bias or not.

On the other hand, the impossible languages in use are based on only a couple of transformations completed only on English. They remain operationally limited to linear and rule-based transformations. Future experiments should perfect such deficiency. To the technical detail of this part, we are unable to create a formal, linguistic definition on impossible languages, as Chomsky's works on generative grammar.

Theoretically, we have not incorporated our findings and foundational construction towards the paradigm shift to other exponents of the Chomskyan theory, such as the Minimalist Program. Theoretical extensions will provide deeper insights for our current perspective towards AI.

\section{Acknowledgments}

This paper is finished for the Generation AI program of Stanford Center at Peking University (SCPKU). We would like to thank Dr. Longlong Ma for his helpful comments and guidance. We would also like to thank Yunwei Pang, a classmate who kindly provided his hardware for our experiments.

\bibliographystyle{acl_natbib}
\bibliography{references}

\appendix

\newpage
\section{Preprocessing}

In our first experiment, after generating the original sentences, we used another python script to transform them into two impossible language datasets with the spaCy library. Sentences are first split, and then dealt with.

In our second experiment, as mentioned before, we employed a BabyLM dataset that includes about 100 million words with the aim to approximate the extent of lexicon that a child before the age of 12 could be exposed to \citep{gilkerson2017}. Another python script is specified to execute the impossible language transformation. We still used the spaCy library, but had to split the complete text into smaller chunks at the beginning. The whole process took several hours in total.

\section{Additional Figures for Experiment 1}

\begin{figure}[!htbp]
    \centering
    \includegraphics[width=0.7\textwidth]{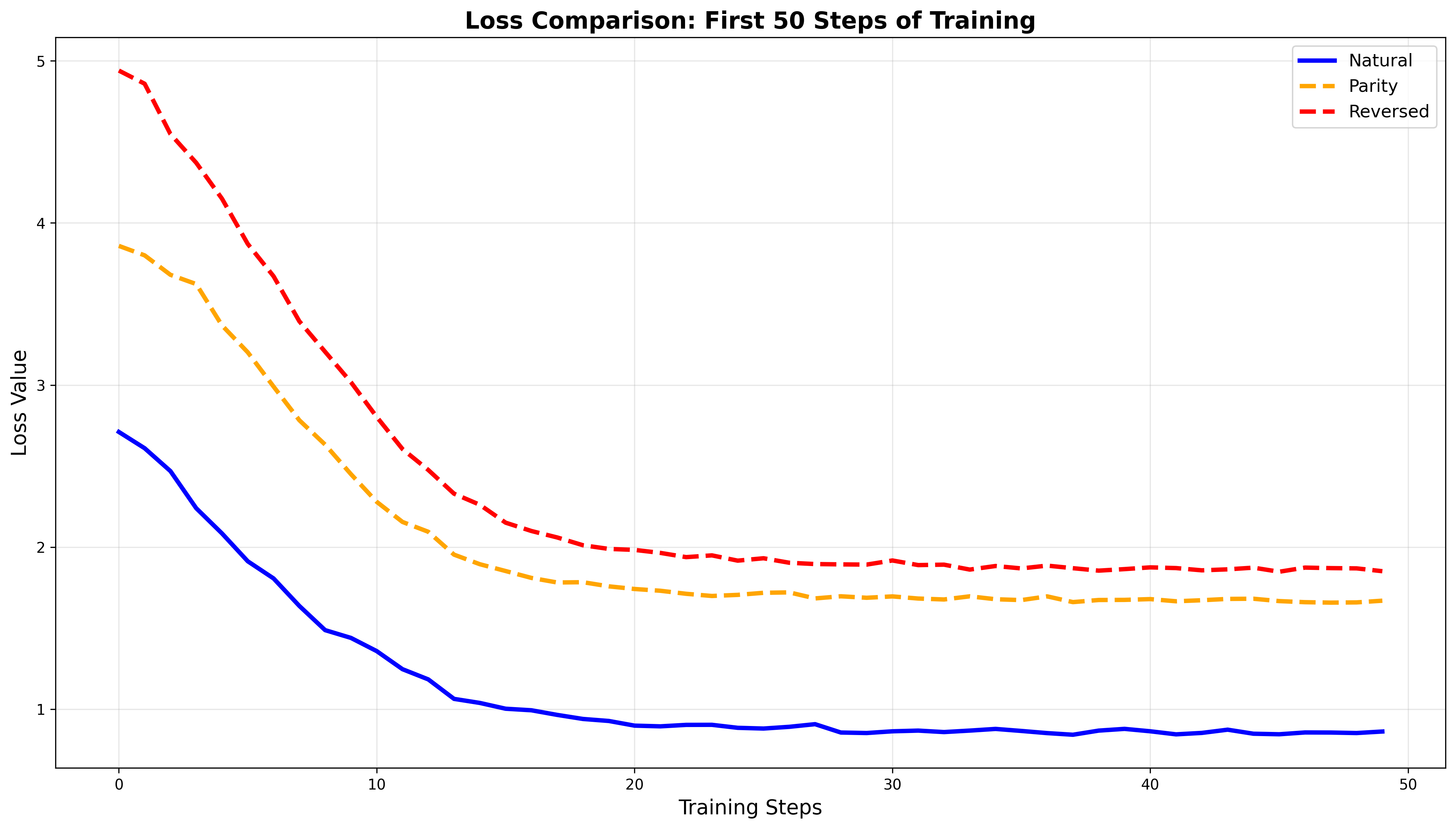}
    \caption{Loss values decline during the first 50 steps in experiment 1}
    \label{fig:app1-first50}
\end{figure}

\begin{figure}[!htbp]
    \centering
    \begin{minipage}{0.48\textwidth}
        \centering
        \includegraphics[width=\textwidth]{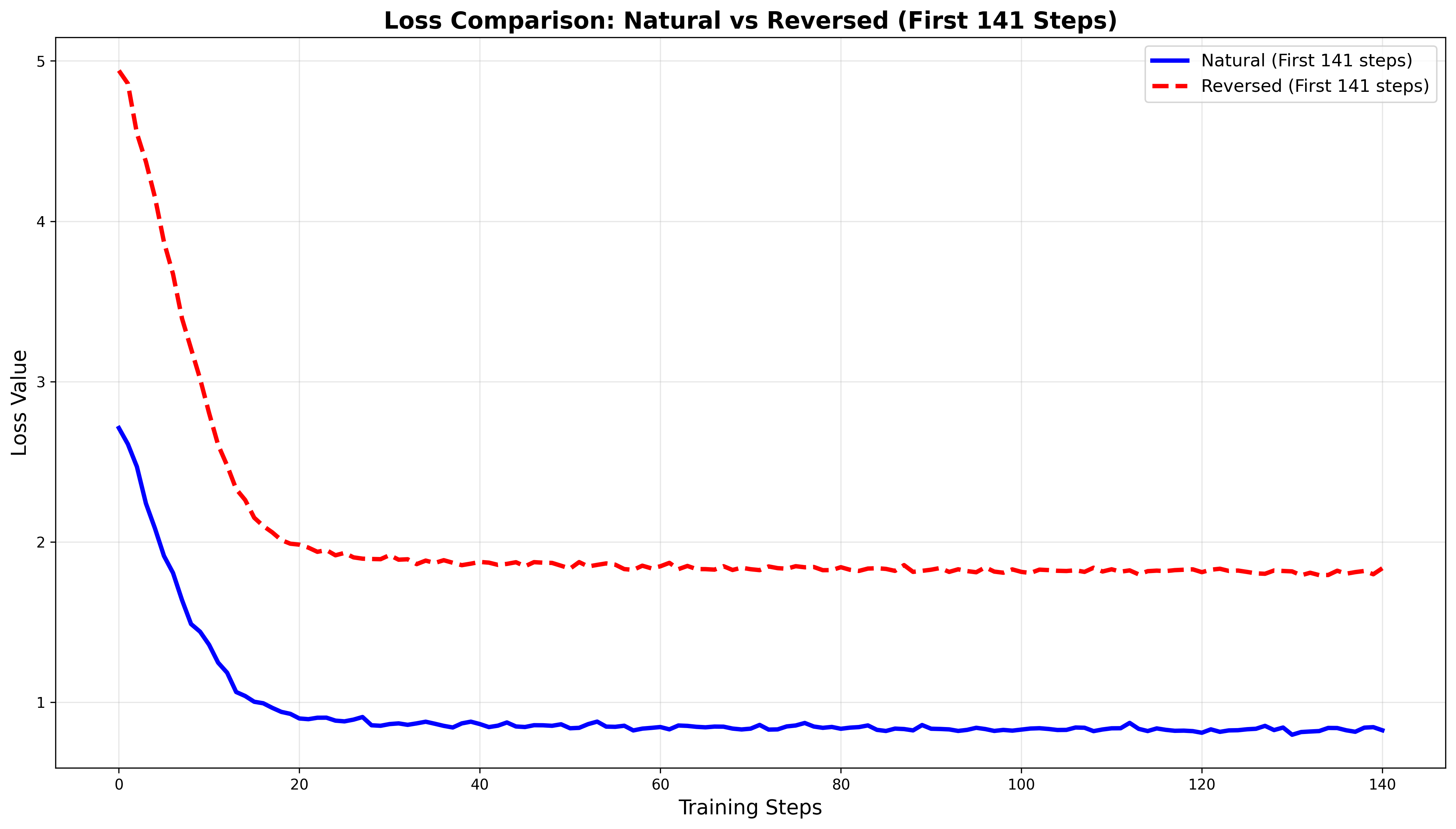}
        \caption{Loss values compared between control group and reversed group in experiment 1}
        \label{fig:app1-reversed}
    \end{minipage}
    \hfill
    \begin{minipage}{0.48\textwidth}
        \centering
        \includegraphics[width=\textwidth]{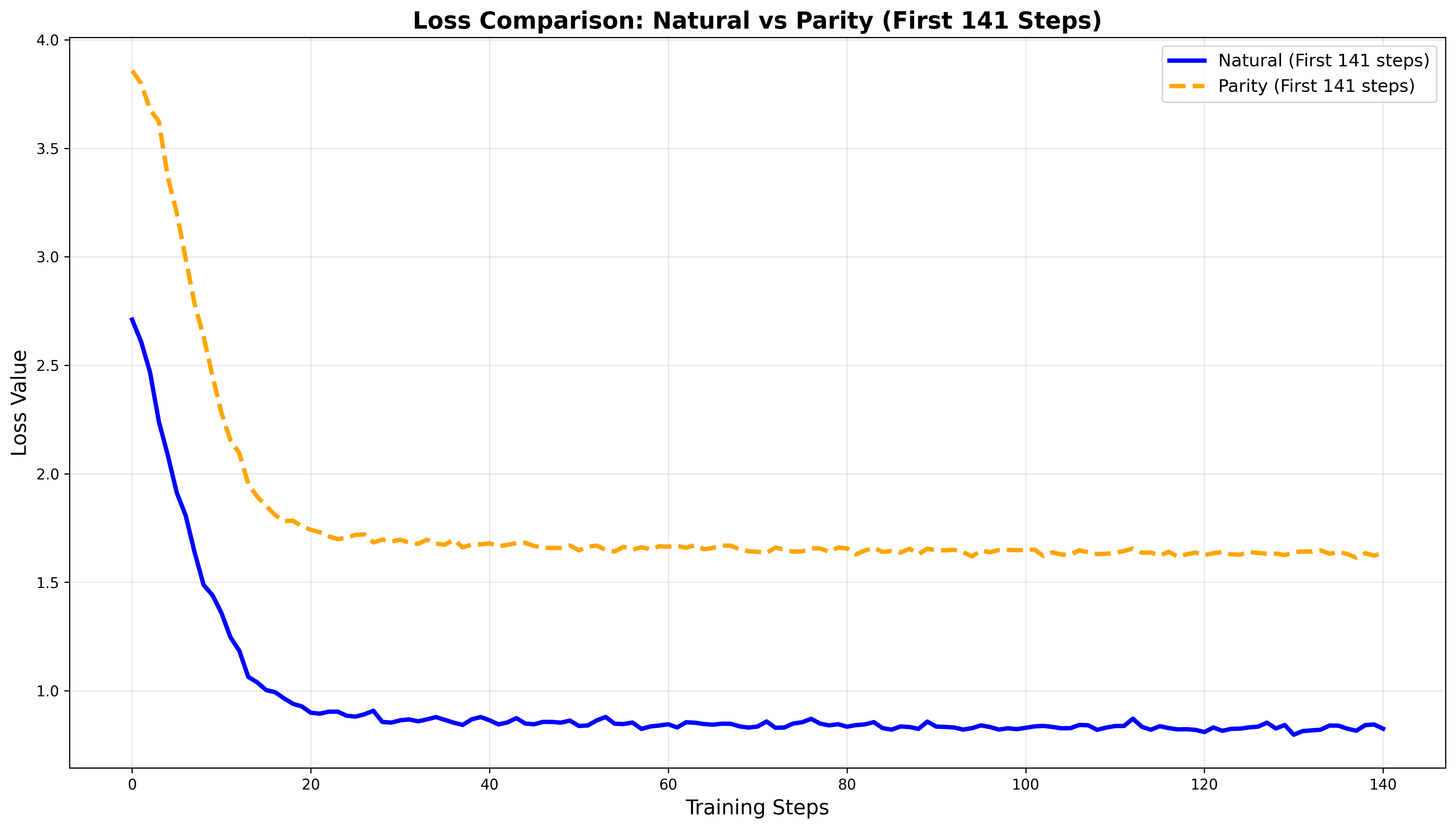}
        \caption{Loss values compared between control group and parity negation group in experiment 1}
        \label{fig:app1-parity}
    \end{minipage}
\end{figure}

\clearpage
\section{Additional Figures for Experiment 2}

\begin{figure}[!htbp]
    \centering
    \includegraphics[width=0.55\textwidth]{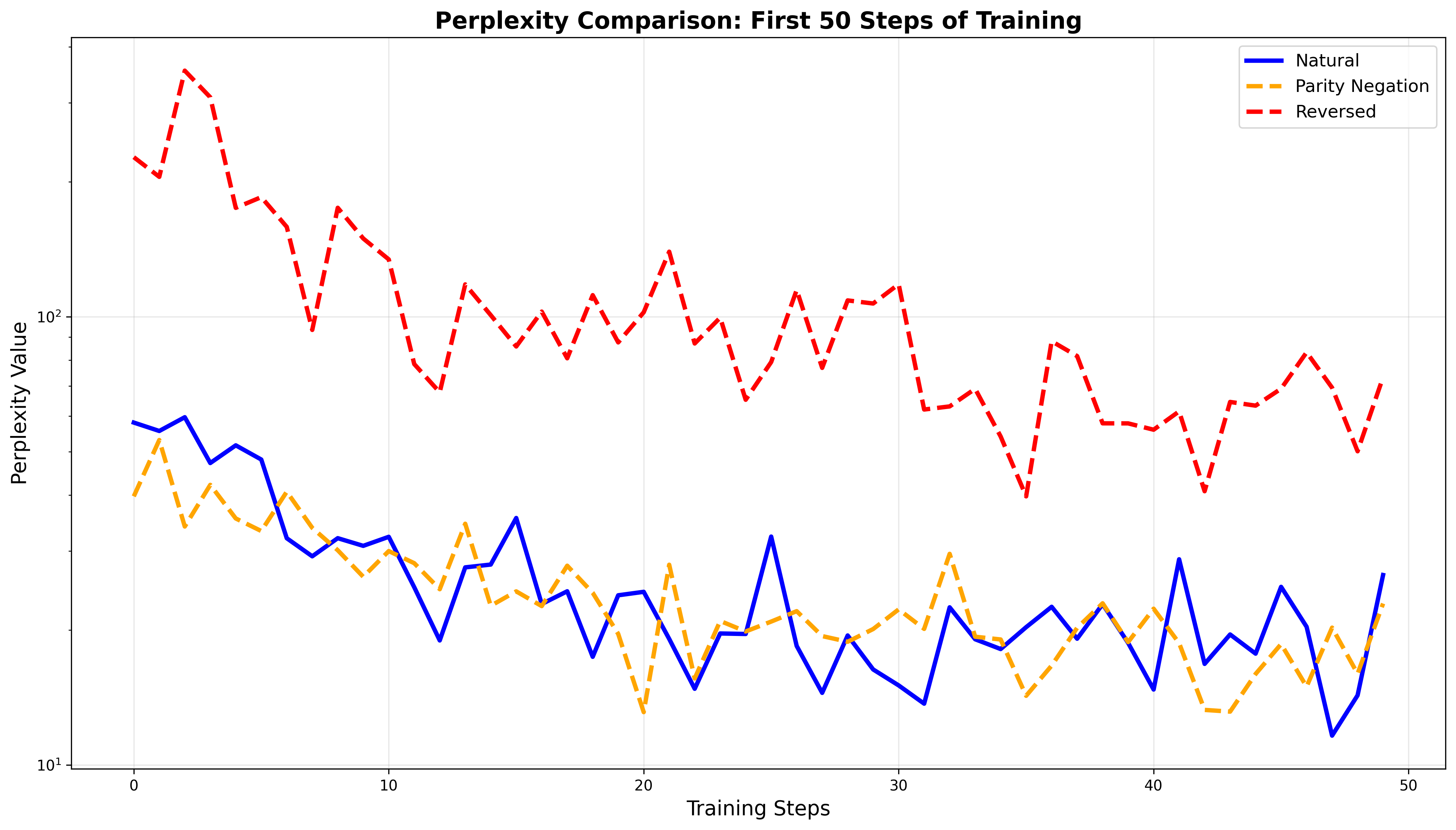}
    \caption{Perplexities decline during the first 50 steps in experiment 2}
    \label{fig:app2-first50}
\end{figure}

\begin{figure}[!htbp]
    \centering
    \begin{minipage}{0.48\textwidth}
        \centering
        \includegraphics[width=\textwidth]{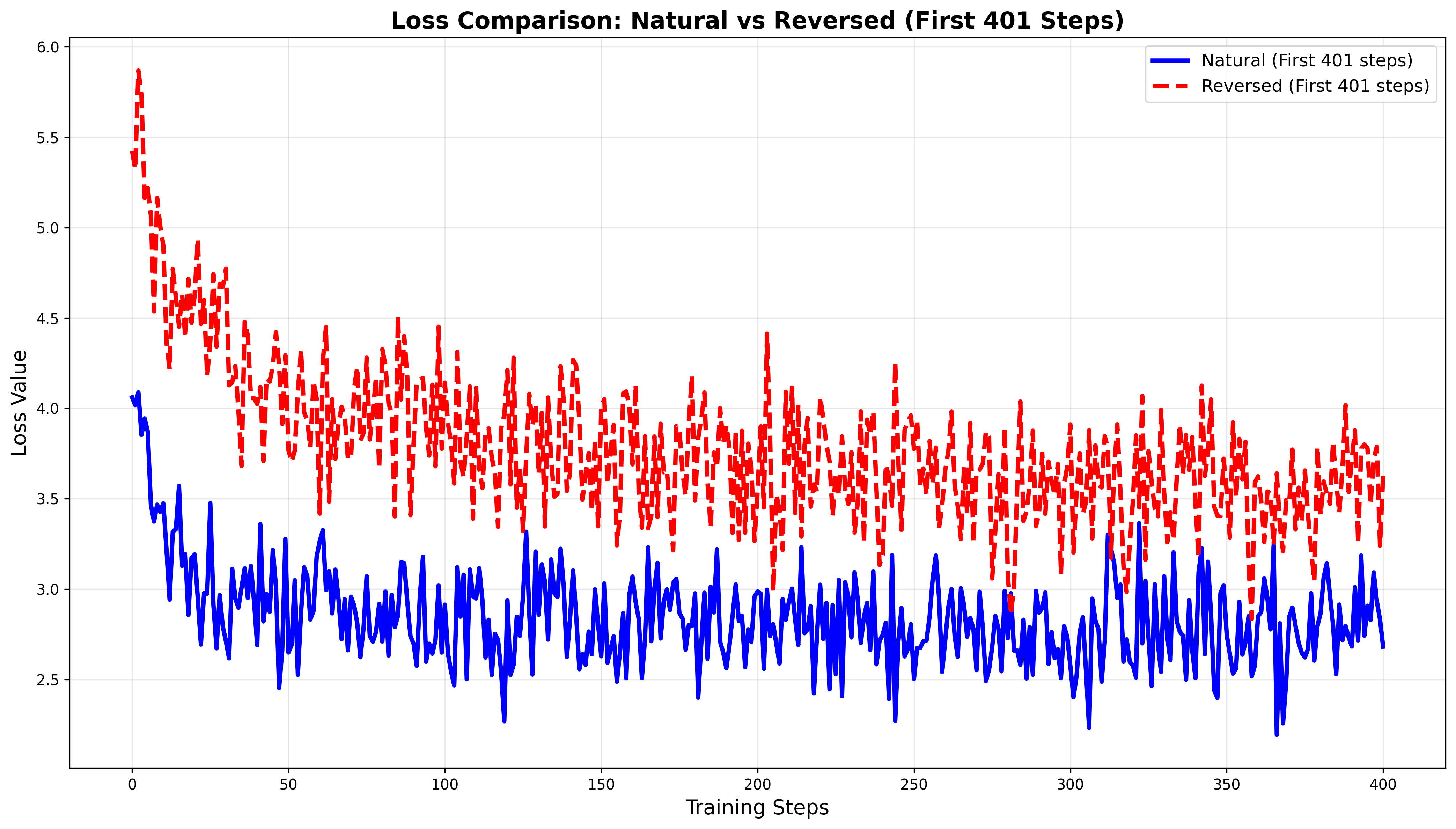}
        \caption{Loss values compared between control group and reversed group in experiment 2}
        \label{fig:app2-reversed}
    \end{minipage}
    \hfill
    \begin{minipage}{0.48\textwidth}
        \centering
        \includegraphics[width=\textwidth]{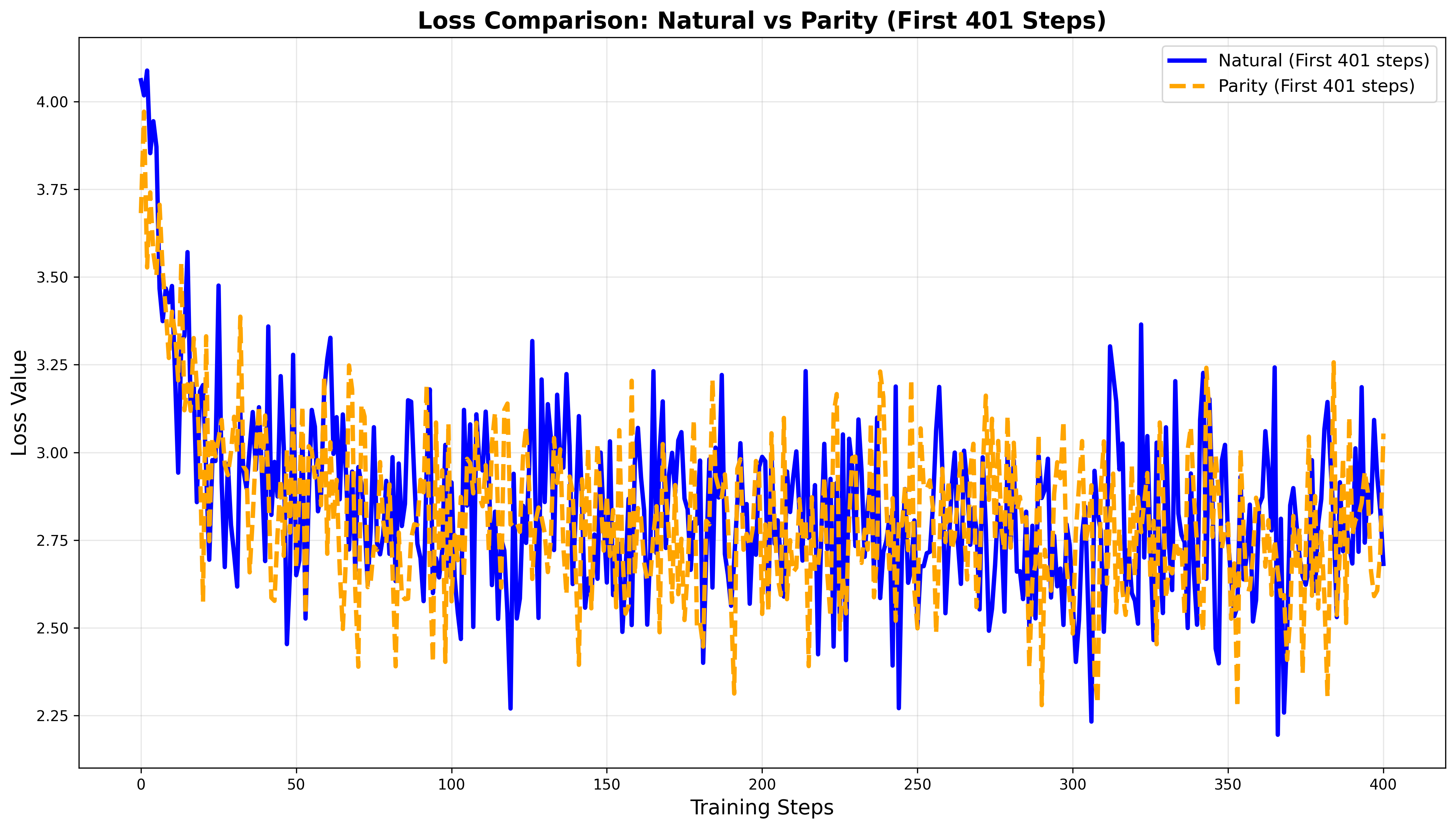}
        \caption{Loss values compared between control group and parity negation group in experiment 2}
        \label{fig:app2-parity}
    \end{minipage}
\end{figure}

\end{document}